\documentclass[10pt,twocolumn,letterpaper]{article}

\usepackage[pagenumbers]{cvpr} % To force page numbers, e.g. for an arXiv version

\usepackage{amsfonts}       % blackboard math symbols
\usepackage{nicefrac}
\usepackage{dsfont}% compact symbols for 1/2, etc.
\usepackage{multirow}
\usepackage{pifont}
\usepackage{cuted}
\usepackage{colortbl}
\usepackage{amsmath}
\usepackage{graphicx}
\usepackage[usenames,dvipsnames]{xcolor}
\definecolor{deformedcolor}{HTML}{AAAA7F} % the #aaaa7f color
\usepackage{tabularx}

\definecolor{best}{RGB}{200, 208, 245}
\definecolor{second}{RGB}{235, 240, 252}
\newcolumntype{Y}{>{\centering\arraybackslash}X}
\usepackage{floatrow}
\usepackage{caption}
\usepackage{setspace} 
\usepackage{adjustbox}
\definecolor{cvprblue}{rgb}{0.21,0.49,0.74}
\newcommand{\name}{Track4World\xspace}
\newfloatcommand{figurebox}{figure}[\nocapbeside][\dimexpr(\textwidth-\columnsep)/2\relax]
\newfloatcommand{tablebox}{table}[\nocapbeside][\dimexpr(\textwidth-\columnsep)/2\relax]
\usepackage[pagebackref,breaklinks,colorlinks,allcolors=cvprblue]{hyperref}

\def\paperID{} % *** Enter the Paper ID here
\def\confName{CVPR}
\def\confYear{2026}

\vspace{-20px}
\title{\name: Feedforward World-centric Dense 3D Tracking of All Pixels}
\author{
Jiahao Lu$^{1}$ \quad
Jiayi Xu$^{1}$ \quad
Wenbo Hu$^{2}\footnotemark[2]~$ \quad
Ruijie Zhu$^{2}$ \quad
Chengfeng Zhao$^{1}$ \quad \\
Sai-Kit Yeung$^{1}$ \quad
Ying Shan$^{2}$ \quad
Yuan Liu$^{1}\footnotemark[2]~$
\\[0.3em]
$^{1}$The Hong Kong University of Science and Technology \\
$^{2}$ARC Lab, Tencent PCG
\\[0.3em]
\small{
Project page:
\href{https://jiah-cloud.github.io/Track4World.github.io/}
{https://jiah-cloud.github.io/Track4World.github.io/}
}
\vspace{-20pt}
}

\begin{document}
\maketitle
\renewcommand{\thefootnote}{\fnsymbol{footnote}}
\footnotetext[2]{Corresponding Authors.}
\begin{strip}\centering
    \vspace{-30px} 
    \captionsetup{type=figure}
    \setlength{\abovecaptionskip}{2pt} 
    \setlength{\belowcaptionskip}{0pt}  
    \includegraphics[width=\textwidth]{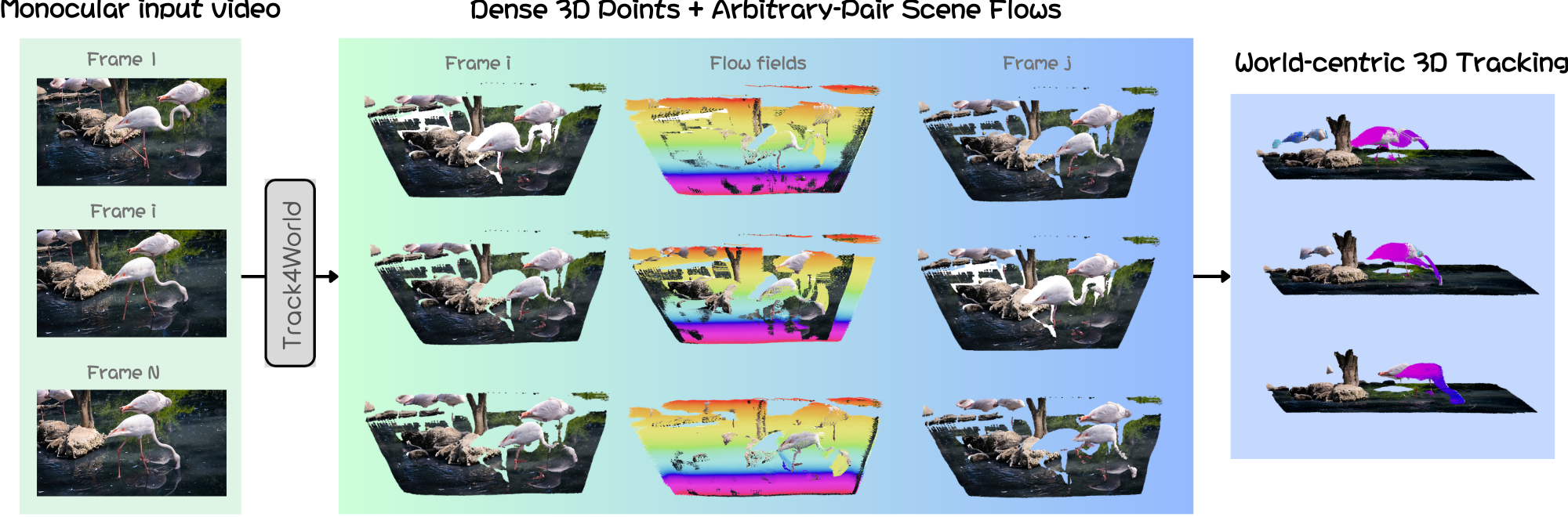}
\caption{\textbf{\name} estimates dense 3D scene flow of every pixel between arbitrary frame pairs from a monocular video in a global feedforward manner, enabling efficient and dense 3D tracking of every pixel in the world-centric coordinate system.
}
    \label{fig:teaser}
    \vspace{-10px}
\end{strip}

\begin{abstract}
Estimating the 3D trajectory of every pixel from a monocular video is crucial and promising for a comprehensive understanding of the 3D dynamics of videos. 
Recent monocular 3D tracking works demonstrate impressive performance, but are limited to either tracking sparse points on the first frame or a slow optimization-based framework for dense tracking. 
In this paper, we propose a feedforward model, called \textbf{\name}, enabling an efficient holistic 3D tracking of every pixel in the world-centric coordinate system.
Built on the global 3D scene representation encoded by a VGGT-style ViT, \name applies a novel 3D correlation scheme to simultaneously estimate the pixel-wise 2D and 3D dense flow between arbitrary frame pairs. The estimated scene flow, along with the reconstructed 3D geometry, enables subsequent efficient 3D tracking of every pixel of this video. 
Extensive experiments on multiple benchmarks demonstrate that our approach consistently outperforms existing methods in 2D/3D flow estimation and 3D tracking, highlighting its robustness and scalability for real-world 4D reconstruction tasks.
\end{abstract}    
\section{Introduction}
\label{sec:intro}

Reconstructing 4D dynamics of videos~\cite{ngo2024delta,xiao2024spatialtracker,ren2023lidar,zhu2024motiongs,cao2025reconstructing,zhang2025tapip3d,lu2025trackingworld,liu2025trace,feng2025st4rtrack,liang2025zero,zhang2025pomato} is a task of great significance, with far-reaching applications~\cite{huang2026pointworld,zhao2024m,gu2025diffusion,burgert2025gowiththeflow} in robotics, animation production, physical law inference, and other fields. To grasp the holistic dynamics of videos, it is essential to predict the dynamic motion of all pixels in 3D space from monocular video inputs. However, this still remains an extremely challenging problem: monocular geometric reconstruction is inherently ill-posed to recover 3D from single-view observations, and further tracking 3D points at different time steps adds an extra layer of complexity.

While recent advances have addressed isolated aspects of this problem, robust and holistic 3D tracking for \textit{every pixel} is still an unsolved problem. Some recent tracking-based methods (e.g., St4RTrack~\cite{feng2025st4rtrack}, SpatialTrackerV2 (STV2)~\cite{xiao2025spatialtrackerv2}, DELTA~\cite{ngo2024delta}), extends feedforward 3D geometry estimation~\cite{wang2025vggt,wang2024dust3r} with temporal association for 3D tracking. Though achieving impressive performance and accuracy, these methods are restricted to 3D tracking of the points on the first frame, failing to capture motions for new pixels in subsequent frames. Another work called TrackingWorld~\cite{lu2025trackingworld} enables dense 3D tracking of all frames by fusing multiple modalities, such as 2D flow~\cite{karaev2024cotracker3}, masks~\cite{yao2025uni4d}, and depth priors~\cite{piccinelli2024unidepth}. While being effective, this pipeline is computationally expensive and limited by the inability to learn joint spatiotemporal priors, often resulting in temporally inconsistent or suboptimal solutions. Following recent trends in feedforward 3D reconstruction methods, such as DUSt3R~\cite{wang2024dust3r}, VGGT~\cite{wang2025vggt}, and Pi3~\cite{wang2025pi}, designing an efficient feedforward model for holistic 3D tracking of every pixel in a monocular video becomes a promising direction.

In this work, we propose \textbf{\name}, a feedforward framework designed to estimate 3D tracking for every pixel of every frame in a monocular video in the world-centric coordinate system. Following the feedforward design trends, \name is built upon the VGGT-like ViT-style 3D geometry reconstruction framework~\cite{wang2025vggt,wang2025pi,lin2025depth} and associates different frames to estimate dense 3D tracking for every pixel in the world-centric coordinate system.

Designing such an association scheme for the 3D tracking of every pixel is computationally expensive. 
A naive solution is to generalize the 3D tracking methods, Spatial Tracker V2 (STV2)~\cite{xiao2025spatialtrackerv2} or DELTA~\cite{ngo2024delta}, to track every pixel.
However, explicitly predicting trajectories for every pixel on all frames is computationally prohibitive. The large number of pixels leads to unaffordable memory and computation consumption during both training and inference, with huge redundant 3D trajectories across different frames.

Instead of directly predicting 3D trajectories of all pixels, \name resorts to predicting the two-frame scene flows as the representation of the holistic 3D video dynamics.
We draw inspiration from efficient pair-wise motion estimation methods, such as St4RTrack~\cite{feng2025st4rtrack} and ZeroMSF~\cite{liang2025zero}, to estimate the dense scene flow between arbitrary frame pairs. Then, the estimated pair-wise motion can be easily utilized to construct the 3D tracking of arbitrary pixels in a video.
By decomposing the continuous tracking problem into pair-wise scene flow estimations, we significantly reduce the computational redundancy, transforming this into a more manageable task. 

However, effectively adapting a VGGT-like ViT-style 3D reconstruction framework for dense scene flow estimation remains an open research problem. A straightforward solution might involve appending a motion decoding head to regress 3D scene flow directly~\cite{feng2025st4rtrack,liang2025zero}. Yet, such implicit regression approaches are typically data-hungry, requiring massive 3D training datasets and heavily parameterized networks. Moreover, they often struggle to capture fine-grained motion details, leading to suboptimal tracking accuracy. 
Conversely, state-of-the-art 3D tracking methods such as STV2~\cite{xiao2025spatialtrackerv2}, and 2D optical flow methods like RAFT~\cite{teed2020raft}, have demonstrated that constructing correlation volumes and iteratively updating the motion field yield significantly more precise results. Building on this insight, we take a step back to the effective and accurate correlation-based estimation scheme to incorporate a novel correlation-based iterative 3D scene flow estimation strategy specifically tailored for dense pixel-level prediction within our feedforward framework. 

Our proposed iterative correlation scheme differs from all existing correlation-based methods to address the specific challenges of dense 3D tracking.

\noindent\textbf{(1) Sparse-to-Dense.} To manage the computational overhead of all-pixel scene flow estimation, we avoid performing iterative correlation updates on the full original resolution. Instead, we operate on a set of sparse anchor points and subsequently recover the dense motion for the entire image via learned upsampling. 

\noindent\textbf{(2) 2D-to-3D Correlation.} We circumvent the expensive 3D spatial correlation (which typically requires $k$-nearest neighbor searches in 3D space followed by cross-attention operations) used in prior works~\cite{xiao2025spatialtrackerv2,zhang2025tapip3d}. We introduce a novel hybrid correlation mechanism that efficiently fuses the 3D geometric feature embeddings from the ViT backbone, utilizing 2D pixel-wise correlations. This design allows us to compute 3D flow updates rapidly without explicitly constructing heavy 3D spatial correlation.

\noindent\textbf{(3) 2D-3D Joint Supervision.} Our 2D-to-3D correlation mechanism inherently supports the dual prediction of 2D and 3D flows. This structural alignment enables a joint supervision strategy that leverages abundant 2D flow datasets to provide auxiliary training signals for the 3D scene flow task. Consequently, we effectively circumvent the severe scarcity of 3D ground-truth annotations, utilizing 2D flow training to significantly enhance the generalization capability of our model.

\noindent\textbf{(4) Global Scene Flow.} Unlike traditional scene flow estimation methods that are limited to adjacent frames, our framework is designed to estimate flows between arbitrary frame pairs within the video sequence, not limiting neighboring frames. By processing the entire video sequence simultaneously, our network leverages global temporal context to resolve local ambiguities, compensating for estimation errors that might occur in isolated frame pairs.

After estimating the accurate pairwise 2D-3D flows for the whole video, \name fuses the 3D trajectories in the global coordinate system and then constructs a holistic 3D tracking of every pixel in the world-centric coordinate system.
Experiment results demonstrate that our approach delivers accurate, flexible, and comprehensive 3D motion estimates. Extensive experiments demonstrate that \name enables robust dense 3D tracking in the world coordinate system, consistently outperforming existing baselines~\cite{liang2025zero,xiao2025spatialtrackerv2,zhang2025pomato}.
\section{Related Work}
\label{sec:rw}

\subsection{Video Geometry Estimation}
Video geometry estimation aims to generate temporally consistent point maps from videos. Early works focus on optimization-based refinement, e.g., RobustCVD~\cite{kopf2021robust} and CasualSAM~\cite{zhang2022structure}, which optimize depth and camera parameters using geometric constraints. Recent differentiable SLAM methods, such as MegaSaM~\cite{li2024megasam}, Uni4D~\cite{yao2025uni4d}, and ViPE~\cite{huang2025vipe}, incorporate depth priors and correspondence maps for video-level point map prediction.
A second line of research explores data-driven approaches~\cite{wang2024dust3r,lu2024align3r,zhang2024monst3r,chen2025easi3r,leroy2024grounding,yang2025fast3r,wang2025vggt,jiang2025geo4d,wang2025continuous,xiong2025human3r,chen2025ttt3r,zhuo2025streaming,li2026unish}, leveraging world-centric geometry for robust, consistent estimation. However, VGGT~\cite{wang2025vggt} shows that camera-centric depth and pose often outperform world-centric representations. Consequently, recent methods, including GeometryCrafter~\cite{xu2025geometrycrafter}, Pi3~\cite{wang2025pi}, and MapAnything~\cite{keetha2025mapanything}, focus on camera-centric predictions. Our approach follows this trend by employing a camera-centric representation (camera-centric point clouds and camera poses), from which world-centric reconstruction can be directly derived, while jointly estimating geometry and motion.

\subsection{Joint Geometry and Motion Estimation}
Joint geometry and motion estimation aims to predict both 3D point clouds and motion trajectories from videos. This is inherently more challenging than geometry estimation alone due to the need for consistent spatial-temporal reasoning.
Existing approaches generally follow three main directions. First, inspired by 2D point tracking~\cite{karaev2024cotracker, karaev2024cotracker3, li2024taptr, harley2025alltracker, qu2024taptrv3, doersch2023tapir}, early methods such as SpatialTracker~\cite{xiao2024spatialtracker} and DELTA~\cite{ngo2024delta} represent points in $(u, v, d)$ coordinates, which require known camera intrinsics and can lead to geometric inaccuracies. Later works, e.g., TAPIP3D~\cite{zhang2025tapip3d} and SpatialTrackerV2~\cite{xiao2025spatialtrackerv2}, track directly in $(x, y, z)$ space for improved stability. They rely on explicit 3D spatial correlations, which typically require performing computationally expensive $k$-nearest neighbor searches in 3D space for each query point, followed by cross-attention with the target points. Second, methods like St4RTrack~\cite{feng2025st4rtrack}, POMATO~\cite{zhang2025pomato}, Stereo4D~\cite{jin2024stereo4d}, and ZeroMSF~\cite{liang2025zero} reconstruct motion by estimating pairwise point maps or scene flow~\cite{teed2021raft,yang2020upgrading,schmid2025ms} between two frames via an implicit regression approach. Third, optimization-based methods such as TrackingWorld~\cite{lu2025trackingworld} combine multiple cues to recover dense 3D tracking for all pixels, but these are computationally expensive and suboptimal for generalization.

Motivated by these methods, we propose a feed-forward, holistic framework for the dense 3D tracking of every pixel. By leveraging a novel 3D correlation scheme, we simultaneously estimate pixel-wise 2D and 3D dense flow between arbitrary frame pairs. The estimated motion, along with the reconstructed 3D geometry, enables the subsequent efficient computation of continuous 3D trajectories for every pixel within the global coordinate system.

\textbf{Concurrent works}. Several concurrent works have explored similar directions in joint geometry and motion estimation~\cite{zhang20254drt,karhade2025any4d,luo20264rc,alumootil2025dept3r,sucar2026vdpm,zhu2026motioncrafterdensegeometrymotion,liu2025trace}. Among these, TraceAnything~\cite{liu2025trace}, Any4D~\cite{karhade2025any4d}, 4RC~\cite{luo20264rc}, DePT3R~\cite{alumootil2025dept3r}, D4RT~\cite{zhang20254drt} and V-DPM~\cite{sucar2026vdpm} also tackle the prediction of motion between timesteps. Track4World introduces two key design choices that distinguish it from these methods. First, we utilize explicit feature correlations to enhance motion prediction accuracy. Second, our novel 2D-lifted correlation architecture alleviates the computational bottleneck inherent in explicit 3D correlation, improving efficiency while enabling the 3D tracking module to benefit from abundant 2D training data.
\section{Method}
\label{sec:method}
\subsection{Overview}

\begin{figure*}[!t]
    \vspace{-1em}
    \begin{center}
        \includegraphics[width=1\textwidth]{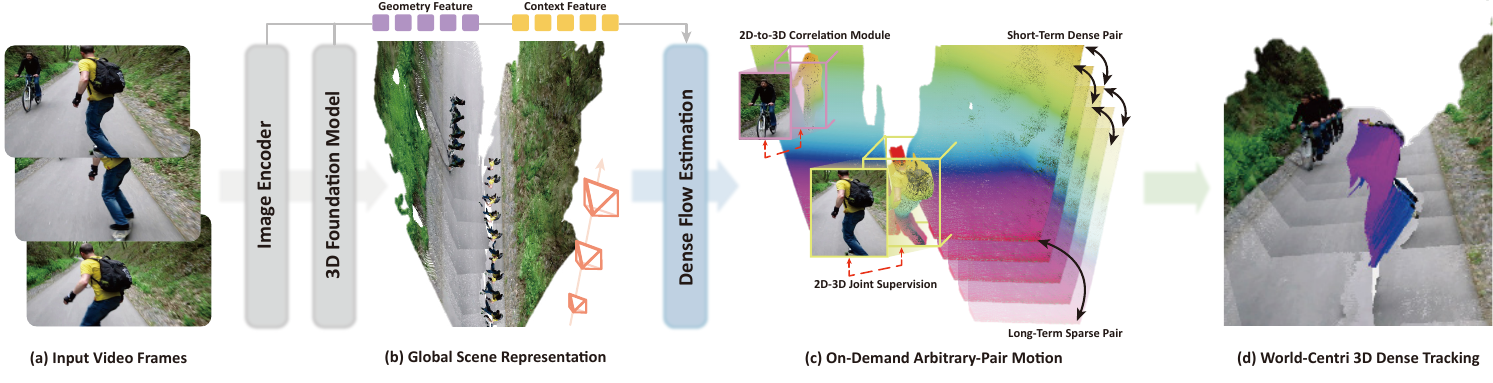}
        \caption{\textbf{Overview}. Given \textbf{(a)} the input video frames, Track4World first extracts \textbf{(b)} global scene representations (geometric embeddings, point clouds, and camera poses). 
        \textbf{(c)} A sparse-to-dense scene flow decoder then predicts 2D-3D joint flows between arbitrary timesteps, which applies a novel 2D-to-3D correlation scheme to improve efficiency and allows 2D-3D joint supervision.
        \textbf{(d)} The pairwise flows are ultimately fused to establish holistic world-centric 3D tracking.
        }
        \label{framework}
    \end{center}
    \vspace{-1em}
\end{figure*}

Given a video sequence of $T$ frames, $\{\mathbf{I}_i \in \mathbb{R}^{H \times W \times 3} \mid i = 1, \ldots, T\}$, \name aims to construct a holistic 3D track for every pixel within a world-centric coordinate system (as illustrated in Fig.~\ref{framework}). To achieve this, rather than treating 3D tracking as an isolated spatial problem, our framework operates through a logical pipeline. First, we extract global scene representations, including geometric features, camera-centric point clouds, and camera poses, using a finetuned geometry encoder initialized from recent state-of-the-art feedforward 3D reconstruction models~\cite{wang2025pi,lin2025depth,wang2025moge1}, such as Pi3 or Depth Anything v3 (DA3), (see supplementary material for details). Second, built upon these global representations, our core scene flow decoder predicts dense 3D scene flow between arbitrary pairs of source and target timesteps $(i, j)$. To reduce computational overhead, this decoder operates in a \textbf{sparse-to-dense} manner. Crucially, it introduces a novel \textbf{2D-to-3D correlation} module that completely avoids expensive 3D spatial correlation, which in turn enables a \textbf{2D-3D joint supervision} strategy to circumvent the severe scarcity of 3D ground-truth annotations. Finally, \name fuses these pairwise 3D scene flows to formulate the ultimate holistic 3D tracks.

\subsection{Scene Flow Decoder}

Building upon the global scene representations (camera-centric pixel-aligned point clouds $\mathbf{P}_i$, camera poses $\mathbf{T}_i$, and geometry features $\mathbf{F}_i$) extracted by the ViT backbone~\cite{wang2025pi,lin2025depth,wang2025moge1}, the scene flow decoder predicts dense 3D scene flow between arbitrary frame pairs $(i, j)$ for the whole video, allowing subsequent 3D tracking in the world-centric coordinate system.

\begin{figure}[!ht]
\vspace{-1em}
    \begin{center}
        \includegraphics[width=1\textwidth]{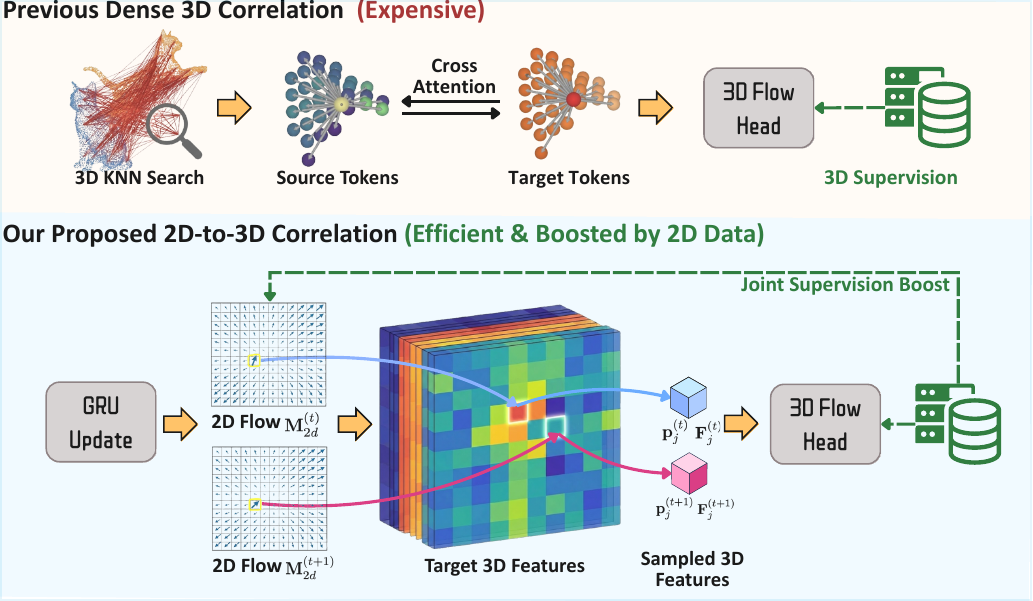}
\caption{\textbf{Comparison of correlation mechanisms.} Prior methods rely on explicit $k$-nearest neighbor searches and cross-attention in 3D space, leading to high computational costs. In contrast, our proposed method anchors 3D updates directly to intermediate image-plane correlations. This design significantly improves computational efficiency and allows the 3D tracking module to be effectively boosted by abundant 2D training data.}
        \label{com3dcorr}
    \end{center}
    \vspace{-1em}
\end{figure}

\subsubsection{Anchor Feature Extraction}
To manage the computational overhead of all-pixel flow estimation, we avoid performing iterative correlation updates at the full original resolution. Instead, we operate on a set of sparse anchor points. We begin with augmenting the geometry tokens $\mathbf{F}_i$ with temporal context using global self-attentions to yield refined features $\hat{\mathbf{F}}_i$. Concurrently, a lightweight context encoder extracts semantic features $\tilde{\mathbf{F}}_i$ directly from the input images. After that, we downsample the point clouds $\mathbf{P}_i$ and both feature maps ($\hat{\mathbf{F}}_i, \tilde{\mathbf{F}}_i$) to $1/8$ resolution, establishing a sparse but contextually rich foundation for the flow estimation, which will be upsampled back to the full resolution flow in the end.

\subsubsection{2D-to-3D Correlation for Flow Estimation}

This section aims to estimate the 3D scene flow on the anchor points using correlation. Previous methods, like STV2~\cite{xiao2025spatialtrackerv2} or TAPIP3D~\cite{zhang2025tapip3d}, demonstrate accurate 3D tracking by the correlation scheme but consume lots of computation in finding $k$-nearest point in 3D space for correlation as shown in Fig.~\ref{com3dcorr} (top), making these methods limited to tracking only sparse 3D points. To scale this for dense 3D flow estimation, we introduce an efficient 2D-to-3D correlation module that elegantly fuses geometric features with image-plane correlations and lifts it for 3D flow estimation. 

In the proposed 2D-3D correlation module, both 2D and 3D flows are estimated with several update iterations to compute the correlations and refine the target positions. In each iteration, the 2D flow and 3D flow are updated sequentially in a coupled manner, as shown in Fig.~\ref{com3dcorr} (bottom). For a given source-target image pair $(i, j)$, the 2D flow is updated first, serving as a reliable spatial basis for the subsequent update of the 3D flow. 
Both 2D flows $\mathbf{M}_{2d}^{(0)}$ and 3D flows $\mathbf{M}_{3d}^{(0)}$ are initialized to 0 everywhere and updated with the following scheme.

\textbf{2D Iterative Correlation}.  We first construct geometric feature correlation volumes $\hat{\mathbf{C}}^{(t)}_{i,j}$ and semantic feature correlation volumes $\tilde{\mathbf{C}}^{(t)}_{i,j}$ between the source position on the source image and the neighborhood positions on the target image using the current 2D flow $\mathbf{M}_{2d}^{(t)}$.    
At the $t$-th iteration, a GRU-based operator updates the hidden context feature from $\hat{\mathbf{F}}^{(t)}_i$ to $\hat{\mathbf{F}}^{(t+1)}_i$, which in turn drives the update of the 2D flow $\mathbf{M}_{2d}^{(t)}$ and visibility confidence $\mathbf{V}^{(t)}$:
\begin{align}
\hat{\mathbf{F}}^{(t+1)}_i 
&= \mathrm{GRU}\big(
    \hat{\mathbf{F}}^{(t)}_i,\,
    \tilde{\mathbf{F}}_i,\,
    \hat{\mathbf{C}}^{(t)}_{i,j},\,
    \tilde{\mathbf{C}}^{(t)}_{i,j},\,
    \mathbf{M}_{2d}^{(t)},\,
    \mathbf{V}^{(t)}
\big), \\
\mathbf{M}_{2d}^{(t+1)} 
&= \mathbf{M}_{2d}^{(t)} 
   + \mathrm{MLP}\big(\hat{\mathbf{F}}^{(t+1)}_i\big), \\
\mathbf{V}^{(t+1)} 
&= \mathbf{V}^{(t)} 
   + \mathrm{MLP}\big(\hat{\mathbf{F}}^{(t+1)}_i\big).
\end{align}
Then, the 2D flow update is used as a basis for 3D flow update as follows.

\textbf{Lifting for 3D Flow Iterative Estimation}. 
With the 2D flow $\mathbf{M}_{2d}^{(t)}$ and $\mathbf{M}_{2d}^{(t+1)}$, we can retrieve the corresponding 2D target positions on timestep $t$ and $t+1$. Then, by interpolating on the extracted global point maps $\mathbf{P}_j$, we can compute 3D coordinates $\mathbf{p}^{(t)}_{j}$ and $\mathbf{p}^{(t+1)}_{j}$ for these two target positions correspondingly. Then, $\mathbf{p}^{(t+1)}_{j}-\mathbf{p}^{(t)}_{j}$ can be an initial estimation of the 3D flow update $\Delta \mathbf{M}_{3d}^{(t)}$. 
However, this only lifts the 2D flow update to 3D without considering any additional 3D geometric correlation. We further fuse all 3D information together using a 3D flow head $\mathcal{H}_{3d}$ to predict the 3D flow update $\Delta \mathbf{M}_{3d}^{(t)}$ by:
\begin{equation}
\begin{split}
\Delta \mathbf{M}^{(t)}_{3d} = \mathcal{H}_{3d}\Big( &
\underbrace{\mathbf{p}^{(t)}_{j}, \mathbf{p}^{(t+1)}_{j}, \mathbf{F}^{(t)}_{j}, \mathbf{F}^{(t+1)}_{j}}_{\text{Lifted Target Samples}}, \\
& \underbrace{\hat{\mathbf{F}}^{(t+1)}_i, \mathbf{F}_i}_{\text{Lifted Source Contexts}}, \underbrace{\mathbf{C}^{(t)}_{3d,i,j}, \mathcal{M}_{3d}^{(t)}}_{\text{Auxiliary Priors}}
\Big),
\end{split}
\end{equation}
where $\mathbf{F}^{(t)}_{j}$ and $\mathbf{F}^{(t+1)}_{j}$ denote the interpolated geometric features of $\mathbf{F}_j$ at the predicted target positions in two consecutive iterations, providing the 3D shape information of these two predicted target positions, ${\mathbf{F}}^{(t+1)}_i, \mathbf{F}_i$ are the source geometric features on the anchor points. We further compute the 3D spatial similarity by correlating the source point cloud warped by the estimated 3D flow $\mathbf{M}^{(t)}_{3d}$ with the target point cloud. The resulting correlation $\mathbf{C}^{(t)}_{3d,i,j}$ between the transformed source points and the target geometry provides informative cues for refining the 3D flow. Additionally, we leverage the historical 3D flow $\mathbf{M}_{3d}^{(t)}$ to construct a trajectory prior $\mathcal{M}_{3d}^{(t)}$, which constrains the smoothness of the flow update.
Then, the 3D flows are updated with $\mathbf{M}_{3d}^{(t+1)} \leftarrow \mathbf{M}_{3d}^{(t)} + \Delta \mathbf{M}_{3d}^{(t)}$.
This 2D-to-3D correlation scheme allows us to accurately and iteratively update both 2D and 3D flows simultaneously while keeping the computation complexity manageable even for all anchor points from all frames.

\subsubsection{Dense Scene Flow Recovery}
\label{dense_recovery}
In the final step, we recover the dense scene flow for the entire image. The low-resolution ($1/8$) flow maps are upsampled to full resolution via a learned pixel-shuffle operation, guided by decoded contextual weights. To maximize the utility of our 2D-lifted architecture, we introduce a hybrid unprojection scheme. Because the intermediate 2D flow captures image-plane dynamics with high precision, we directly combine it with the Z-axis displacement from the 3D scene flow. Using camera intrinsics derived from the predicted point clouds, we project this combined motion into $(x, y, z)$ space to produce the final refined 3D scene flow. The benefits of this hybrid design are analyzed in Tab.~\ref{table:ablation}.

\subsection{Discussion}
\label{temporaltracking}

\noindent\textbf{Computational Efficiency.}
Prior works~\cite{xiao2025spatialtrackerv2,zhang2025tapip3d} heavily rely on explicit 3D spatial correlations, which typically require a computationally expensive $k$-nearest neighbor ($k$-NN) search in 3D space for each query point, followed by a cross-attention mechanism with the retrieved target points. In contrast, our 3D flow estimation is fundamentally driven by the proposed 2D-lifted correlation module. This design completely bypasses both the 3D $k$-NN search and the heavy cross-attention operations, significantly alleviating the computational burden. Specifically, while traditional 3D point-based matching incurs a theoretical complexity of at least $\mathcal{O}(N \log N + N \cdot k)$ (or $\mathcal{O}(N^2)$ for dense global attention) for $N$ points, our warp-sampling mechanism operates with a strict $\mathcal{O}(N)$ complexity via direct coordinate-based lookups from the 2D flow.

\noindent\textbf{2D-3D Joint Supervision.} This structural alignment naturally supports dual supervision. Unlike prior arts~\cite{xiao2025spatialtrackerv2,zhang2025tapip3d} that define correlations strictly in 3D space, our architecture anchors the 3D updates directly to image-plane correlations. This allows us to apply auxiliary training signals from abundant 2D datasets directly to the intermediate 2D motion priors. By leveraging massive 2D data to guide the lifting process, we effectively circumvent the severe scarcity of 3D ground-truth annotations and significantly enhance the generalization capability of our model.

\noindent\textbf{Global Trajectory Inference.}
Leveraging the predicted arbitrary-pair dense flows, our framework seamlessly generalizes to global 3D tracking. For long-range point tracking, we infer flows from a reference frame to all subsequent frames. This process is refined by a \textit{temporal aggregator} that applies attention over concatenated flow features to enforce temporal consistency. Conversely, for dense tracking of every pixel, we compute and chain consecutive frame-to-frame flows, yielding continuous 3D trajectories in the global coordinate system.

\noindent\textbf{Short-to-Long-Term Supervision.}
During training, we implement a variable-stride sampling strategy: the model is simultaneously supervised by dense flow ground truth over varying temporal intervals and sparse trajectory annotations across the entire sequence. By optimizing these short- and long-term objectives jointly, the network learns to reconcile local motion precision with global trajectory consistency, constructing a holistic 3D track for every pixel.
\section{Experiments}
\label{sec:exper}

\textbf{Implementation details}. Notably, we employ tailored training strategies to fine-tune the base geometric model, leading to refined point map and camera pose estimations as well. Unless specified, all reported results use DA3~\cite{lin2025depth} as the default backbone initialization. Implementation details about specific fine-tuning strategies are provided in the supplementary materials. 
\begin{table*}[!t]
  \begin{center}
    \footnotesize
    \setlength\tabcolsep{1.5pt}
    \vspace{-0.6em}
    \caption{\textbf{Evaluation on scene and optical flow estimation.} We evaluate our model on one in-domain dataset: Kubirc-3D val~\cite{greff2022kubric} and two out-of-domain datasets: KITTI~\cite{geiger2013vision}, BlinkVision~\cite{li2024blinkvision}. \textbf{Best} results are highlighted in \colorbox{best}{darker blue}, and \textbf{second best} in \colorbox{second}{lighter blue}.}
    \label{table:flow}
    \begin{adjustbox}{width=\textwidth}{\begin{tabular}{l|cccccccc|cccccccccc}
      \toprule
       \multirow{3}{*}{Method} & \multicolumn{16}{c}{In domain}\\
       \cline{2-17}
   &\multicolumn{8}{c|}{Kubric-3D val~\cite{greff2022kubric} (short)} &    \multicolumn{8}{c}{Kubric-3D val (long)}  \\

       & Abs Rel $\downarrow$& $\delta<1.25 \uparrow$    & EPE3D$\downarrow$ &AccS $\uparrow$ &AccR $\uparrow$ &EPE2D$\downarrow$ &AccS$_{2D}$$\uparrow$  &AccR$_{2D}$$\uparrow$ & Abs Rel $\downarrow$& $\delta<1.25 \uparrow$    & EPE3D$\downarrow$ &AccS $\uparrow$ &AccR $\uparrow$ &EPE2D$\downarrow$ &AccS$_{2D}$$\uparrow$  &AccR$_{2D}$$\uparrow$ \\
       \midrule
       RAFT~\cite{teed2020raft}  & /& /& /&  /& /& \cellcolor{second}{6.7974} & 0.7442 & 0.9018 & /& /& /& /&  /& \cellcolor{second}{51.9034} & 0.5183 & 0.7042\\
GMFlowNet~\cite{zhao2022global}   & /& /& /& /&  /& 7.0390 & 0.7619 & \cellcolor{second}{0.9103} & /& /& /&  /& /& 51.9049 & 0.5309 & \cellcolor{second}{0.7201}\\
SEA-RAFT~\cite{wang2024sea} & /& /& /&  /& /& 9.5794 & \cellcolor{second}{0.7720} & 0.8947 & /& /& /& /&  /& 58.7148 & \cellcolor{second}{0.5596} & 0.6825\\
       RAFT-3D~\cite{teed2021raft} & 0.0649 & 0.9344 & 0.6170 & 0.0015 & 0.0078 & 40.4480 & 0.0002 & 0.0015 & 0.1245 & 0.8422 & 1.5652 & 0.0001 & 0.0010
& 83.6966 & 0.0004 & 0.0040\\
OpticalExpansion~\cite{yang2020upgrading} & 0.2170 & 0.6266 & \cellcolor{second}{0.2093} & \cellcolor{second}{0.2890} & \cellcolor{second}{0.4760}  & 19.6471 & 0.1183 & 0.3316 & 0.2177 & 0.6316 & \cellcolor{second}{0.7037} & \cellcolor{second}{0.1062} & \cellcolor{second}{0.1903} & 68.6562 & 0.0255 & 0.0874 \\
POMATO~\cite{zhang2025pomato} & 0.1525 & 0.8329 & 0.9672 & 0.0566 & 0.1696  & / & / & / & 0.1761 & 0.7760 & 1.6925 & 0.0148 & 0.0564  & / & / & /\\
ZeroMSF~\cite{liang2025zero} & 0.0860 & 0.9196 & 0.3528 & 0.1867 & 0.3413 & / & / & / & 0.1208 & 0.8609 & 1.2182 & 0.0475 & 0.0895 & / & / & /\\
Any4D~\cite{karhade2025any4d} & \cellcolor{second}{0.0585}&\cellcolor{second}{0.9547}&0.3908&0.1610&0.2893&/&/&/&\cellcolor{second}{0.1017}&\cellcolor{second}{0.8770}&1.2442& 0.0429&0.0855&/&/&/ \\
V-DPM~\cite{sucar2026vdpm}&0.0716&0.9010&0.4087&0.1442&0.2491& / & / & /&0.1155&0.8205&1.2620&0.0407&0.0803& / & / & /\\

\midrule
\textbf{Ours} & \cellcolor{best}{0.0344} & \cellcolor{best}{0.9719} & \cellcolor{best}{0.1537} & \cellcolor{best}{0.5494} & \cellcolor{best}{0.7460}  & \cellcolor{best}{1.8685} & \cellcolor{best}{0.8086} & \cellcolor{best}{0.9309} & \cellcolor{best}{0.0472} & \cellcolor{best}{0.9371} & \cellcolor{best}{0.4808} & \cellcolor{best}{0.3247} & \cellcolor{best}{0.5491}  & \cellcolor{best}{15.0906} & \cellcolor{best}{0.6134} & \cellcolor{best}{0.7711} \\
\midrule
\multirow{3}{*}{Method} &\multicolumn{16}{c}{Out of domain} \\
\cmidrule{2-17}
 & \multicolumn{8}{c|}{KITTI~\cite{geiger2013vision}} & \multicolumn{8}{c}{BlinkVision~\cite{li2024blinkvision}}  \\
 & Abs Rel $\downarrow$ & $\delta<1.25 \uparrow$ & EPE3D $\downarrow$ & AccS $\uparrow$ & AccR $\uparrow$ & EPE2D $\downarrow$ & AccS$_{2D}$ $\uparrow$ & AccR$_{2D}$ $\uparrow$
 & Abs Rel $\downarrow$ & $\delta<1.25 \uparrow$ & EPE3D $\downarrow$ & AccS $\uparrow$ & AccR $\uparrow$  & EPE2D $\downarrow$ & AccS$_{2D}$ $\uparrow$ & AccR$_{2D}$ $\uparrow$ \\
\midrule
RAFT~\cite{teed2020raft} & /& /& /&  /& /& 5.4150 & 0.6271 & 0.8068 & /& /& /& /&  /& 14.1255 & 0.5037 & 0.6953\\
GMFlowNet~\cite{zhao2022global} & /&  /& /& /& /& \cellcolor{second}{4.6977} & 0.6432 & 0.8241 & /& /& /& /& /& \cellcolor{second}{12.0176} & \cellcolor{second}{0.5281} & 0.7170\\
SEA-RAFT~\cite{wang2024sea} & /&  /& /& /& /& 4.8863 & \cellcolor{second}{0.6654} & \cellcolor{second}{0.8297} & /& /& /& /& /& 20.9160 & \cellcolor{best}{0.5697} & \cellcolor{second}{0.7186}\\
RAFT-3D~\cite{teed2021raft} & 0.1619 & \cellcolor{second}{0.8413} & 0.3837 & 0.0118 & 0.0678 & 54.0938 & 0.0001 & 0.0007 & \cellcolor{second}{0.1426} & \cellcolor{second}{0.8455} & 0.6690 & 0.0454 & 0.1280  & 85.4975 & 0.0018 & 0.0121 \\
OpticalExpansion~\cite{yang2020upgrading} & 0.2764 & 0.4302 & 0.2419 & 0.1553 & 0.2612  & 8.8808 & 0.5446 & 0.7326 & 0.3372 & 0.4099 & 0.4406 & \cellcolor{second}{0.2091} & \cellcolor{second}{0.3116}  & 20.2384 & 0.4122 & 0.6139 \\
POMATO~\cite{zhang2025pomato} & 0.2752 & 0.4359 & 0.2602 & 0.1127 & 0.2156  & / & / & / & 0.2089 & 0.6569 & 0.4038 & 0.1522 & 0.2870  & / & / & / \\
ZeroMSF~\cite{liang2025zero} & 0.2064 & 0.5913 & \cellcolor{second}{0.1823} & \cellcolor{second}{0.1695} & \cellcolor{second}{0.3481} & / & / & / & 0.1934 & 0.6620 & \cellcolor{second}{0.3937} & 0.1913 & 0.2991 &  / & / & / \\
Any4D~\cite{karhade2025any4d} & 0.2398
&0.4974&0.1856&0.1429&0.2931&/&/&/&0.2218&0.6125&0.9238&0.1242&0.1818&/&/&/ \\
V-DPM~\cite{sucar2026vdpm}&\cellcolor{second}{0.1469}&0.7981&0.4462&0.1180&0.1608& / & / & /&0.2117&0.6449&1.1476&0.1079&0.1547& / & / & /\\
\midrule
\textbf{Ours} & \cellcolor{best}{0.0707} & \cellcolor{best}{0.9570} & \cellcolor{best}{0.0742} & \cellcolor{best}{0.6929} & \cellcolor{best}{0.8238}  & \cellcolor{best}{2.5722} & \cellcolor{best}{0.6849} & \cellcolor{best}{0.8769} & \cellcolor{best}{0.0371} & \cellcolor{best}{0.9768} & \cellcolor{best}{0.1135} & \cellcolor{best}{0.5091} & \cellcolor{best}{0.7144}  & \cellcolor{best}{7.5632} & 0.5131 & \cellcolor{best}{0.7424} \\
\bottomrule
    \end{tabular}}
    \end{adjustbox}
     \vspace{-2.2em}
  \end{center}
\end{table*}
\noindent\textbf{Experiment overview}. To comprehensively evaluate our framework, we conduct experiments across a variety of tasks, including 2D/3D flow estimation, 2D/3D tracking, point map prediction, and camera pose estimation. Furthermore, we present qualitative results on diverse in-the-wild videos and conduct extensive ablation studies to validate the effectiveness of each proposed module.

\subsection{Scene and Optical Flow Estimation}
\textbf{Evaluation datasets.} 
We evaluate our model on one in-domain dataset, Kubric-3D val~\cite{greff2022kubric}, 
and two out-of-domain datasets, KITTI~\cite{geiger2013vision} and BlinkVision~\cite{li2024blinkvision}. 
All three datasets provide ground-truth optical flow and scene flow annotations. 
For the Kubric-3D dataset, we sample two evaluation settings: 
a \textit{short-range} setting where the source and reference frames are 4 frames apart, 
and a \textit{long-range} setting where they are 16 frames apart, 
to assess flow estimation performance under different temporal gaps. For the BlinkVision~\cite{li2024blinkvision} dataset, the source–target frame pair is randomly sampled.

\noindent\textbf{Evaluation metrics.} For scene flow estimation, we follow previous works~\cite{liang2025zero,gu2019hplflownet,liu2019flownet3d,teed2021raft} and report the standard metrics: End Point Error (EPE3D), Accuracy Strict (AccS), and Accuracy Relax (AccR). Additionally, we evaluate the quality of the previous-frame point cloud and the transformed point cloud obtained by applying the predicted scene flow. We adopt the absolute relative error (Abs Rel) and the percentage of inlier points ($\delta < 1.25$) for this assessment.
To ensure a fair comparison, both predicted point clouds are aligned to the ground truth by optimizing a shared scale factor and translation. This scale factor is further applied to align the predicted scene flow before computing EPE3D, AccS, and AccR.
This alignment procedure follows the method described in MoGe~\cite{wang2025moge1}. For optical flow estimation, we follow~\cite{teed2020raft,zhao2022global,wang2024sea} to report 3 metrics, EPE2D, AccS$_{2D}$, AccR$_{2D}$.

\noindent\textbf{Comparison with existing methods.}
We compare our method against several representative approaches: three optical flow methods (RAFT~\cite{teed2020raft}, GMFlowNet~\cite{zhao2022global}, and SEA-RAFT~\cite{wang2024sea}), two scene flow methods (RAFT-3D~\cite{teed2021raft} and OpticalExpansion~\cite{yang2020upgrading}), two joint geometry and scene flow methods (POMATO~\cite{zhang2025pomato} and ZeroMSF~\cite{liang2025zero}), as well as concurrent works Any4D~\cite{karhade2025any4d} and V-DPM~\cite{sucar2026vdpm}. For RAFT-3D, we provide the depth and camera parameters predicted by VGGT~\cite{wang2025vggt} as input, while OpticalExpansion uses the point clouds predicted by POMATO. Tab.~\ref{table:flow} presents a quantitative comparison, demonstrating that our method consistently outperforms these prior works across all four datasets. This validates the effectiveness of the proposed 2D-3D correlation for the flow estimation task.

\begin{table*}[!t]
\centering
\footnotesize
\renewcommand{\arraystretch}{1.15} 
\setlength{\tabcolsep}{4pt}      

\begin{tabularx}{\textwidth}{l Y Y Y Y Y Y Y Y Y Y}
\toprule
\multirow{2}{*}{Method} 
& \multicolumn{2}{c}{PointOdyssey~\cite{zheng2023pointodyssey}} 
& \multicolumn{2}{c}{ADT~\cite{pan2023aria}} 
& \multicolumn{2}{c}{PStudio~\cite{joo2015panoptic}} 
& \multicolumn{2}{c}{DriveTrack~\cite{balasingam2024drivetrack}} 
& \multicolumn{2}{c}{\textbf{Avg.}} \\

\cmidrule(lr){2-3} \cmidrule(lr){4-5} \cmidrule(lr){6-7} \cmidrule(lr){8-9} \cmidrule(lr){10-11}
 & L-16 & L-50 & L-16& L-50 & L-16 & L-50 & L-16 & L-50 & L-16& L-50 \\
\midrule

% --- Section 1: Camera Coordinate ---
\multicolumn{11}{l}{\textit{Camera coordinate 3D tracking}} \\
\midrule
SpatialTracker$^*$~\cite{xiao2024spatialtracker} & 0.3116 & 0.2977 & 0.4962 & 0.4692 & 0.5390 & 0.4991 & 0.2529 & 0.2502 & 0.3999 & 0.3791 \\
DELTA$^*$~\cite{ngo2024delta} & 0.3529 & 0.3412 & 0.5116 & 0.4952 & \cellcolor{second}{0.5922} & \cellcolor{best}{0.5533} & 0.2704 & 0.2701 & 0.4317 & 0.4150 \\
STV2$^\dagger$~\cite{xiao2025spatialtrackerv2} & 0.1864 & 0.1785 & 0.2400 & 0.2330 & 0.3784 & 0.3690 & 0.1711 & 0.1725 & 0.2400 & 0.2383 \\
\midrule
MASt3R~\cite{leroy2024grounding} & 0.3546 & 0.3253 & 0.3368 & 0.3029 & 0.3293 & 0.2956 & 0.2767 & 0.2559 & 0.3244 & 0.2949 \\
MonST3R~\cite{zhang2024monst3r} & 0.3912 & 0.3860 & 0.3694 & 0.3429 & 0.3511 & 0.3381 & 0.3056 & 0.2787 & 0.3543 & 0.3364 \\
POMATO~\cite{zhang2025pomato} & \cellcolor{second}{0.4816} & \cellcolor{second}{0.4623} & 0.5338 & \cellcolor{second}{0.5299} & 0.5163 & 0.4726 & 0.4237 & 0.4329 & \cellcolor{second}{0.4888} & \cellcolor{second}{0.4744} \\
ZeroMSF~\cite{liang2025zero} & 0.4214 & 0.3887 & \cellcolor{second}{0.5382} & 0.4635 & 0.5083 & 0.4524 & \cellcolor{second}{0.4448} & \cellcolor{second}{0.4513} & 0.4782 & 0.4390 \\

% Ours - Best
\textbf{Ours} & \cellcolor{best}{0.5397} & \cellcolor{best}{0.5268} & \cellcolor{best}{0.6501} & \cellcolor{best}{0.6091} & \cellcolor{best}{0.5948} & \cellcolor{second}{0.5423} & \cellcolor{best}{0.5003} & \cellcolor{best}{0.5092} & \cellcolor{best}{0.5712} & \cellcolor{best}{0.5469} \\

\midrule
% --- Section 2: World Coordinate ---
\multicolumn{11}{l}{\textit{World coordinate 3D tracking}} \\
\midrule
STV2$^\dagger$~\cite{xiao2025spatialtrackerv2} & 0.1925 & 0.1763 & 0.2456 & 0.2163 & 0.3790 & 0.3689 & 0.1711 & 0.1725 & 0.2470 & 0.2335 \\
POMATO$^\ddagger$~\cite{zhang2025pomato} & 0.4425 & 0.3905 & 0.3611 & 0.3548 & 0.5166 & 0.4713 & 0.4227 & 0.4210 & 0.4357 & 0.4094 \\
ZeroMSF$^\ddagger$~\cite{liang2025zero} & 0.4053 & 0.3505 & 0.4530 & 0.3563 & 0.4828 & 0.4386 & 0.4474 & 0.4382 & 0.4471 & 0.3959 \\
Any4D~\cite{karhade2025any4d} &0.4769& 0.4174&0.4460&0.3717&0.5707&0.5066&\cellcolor{best}{0.5235}&\cellcolor{second}{0.5079}&0.5043&0.4509\\
V-DPM~\cite{sucar2026vdpm}&\cellcolor{second}{0.4848}&\cellcolor{second}{0.4233}&\cellcolor{second}{0.4783}&\cellcolor{second}{0.3759}&\cellcolor{best}{0.6084}&\cellcolor{best}{0.5795}&0.4854&0.4817 &\cellcolor{second}{0.5142}&\cellcolor{second}{0.4668}\\
% Ours -\\
\textbf{Ours} & \cellcolor{best}{0.5345} & \cellcolor{best}{0.5162} & \cellcolor{best}{0.6250} & \cellcolor{best}{0.5622} & \cellcolor{second}{0.5946} & \cellcolor{second}{0.5422} & \cellcolor{second}{0.5003} & \cellcolor{best}{0.5087} & \cellcolor{best}{0.5636} & \cellcolor{best}{0.5323} \\

\bottomrule
\end{tabularx}
\caption{\textbf{3D tracking estimation.} We report the APD metric to evaluate 3D point tracking on the PointOdyssey~\cite{zheng2023pointodyssey}, ADT~\cite{pan2023aria}, PStudio~\cite{joo2015panoptic}, and DriveTrack~\cite{balasingam2024drivetrack} datasets. L-16 and L-50 indicate tracking within the temporal length of 16 and 50 frames, respectively. \textbf{*} indicates that ground-truth camera intrinsics are used. $\dagger$ indicates that bundle adjustment is not used. $^\ddagger$ denotes methods using camera poses estimated by VGGT~\cite{wang2025vggt}.}
\label{tab:3dtracking}
\vspace{-0.6em}
\end{table*}

\subsection{3D Tracking Estimation}
\textbf{Evaluation datasets.} Following prior works~\cite{ngo2024delta,xiao2024spatialtracker,zhang2025pomato,xiao2025spatialtrackerv2}, we evaluate our method on ADT~\cite{pan2023aria}, PStudio~\cite{joo2015panoptic}, and DriveTrack~\cite{balasingam2024drivetrack} benchmarks from the TAPVid-3D~\cite{koppula2024tapvid} dataset, as well as the validation set of the PointOdyssey~\cite{zheng2023pointodyssey} dataset. Consistent with POMATO~\cite{zhang2025pomato}, we reformulate the datasets by projecting all query points within a temporal window to the first frame. The window sizes are 16 and 50 in the evaluation.

\noindent\textbf{Evaluation metrics.} We use the Average Percent Deviation (APD) metric, which measures the percentage of points whose predicted positions fall within a specified threshold relative to the ground-truth depth. APD provides a direct and quantitative assessment of the tracking accuracy.

\noindent\textbf{Comparison with existing methods.} We compare our method with representative 3D tracking and geometry/scene flow approaches. The baselines include 3D trackers SpatialTracker~\cite{xiao2024spatialtracker} and DELTA~\cite{ngo2024delta} (both using ground-truth intrinsics), as well as STV2~\cite{xiao2025spatialtrackerv2}. We further compare against geometry and scene flow methods MASt3R~\cite{leroy2024grounding}, MonST3R~\cite{zhang2024monst3r}, POMATO~\cite{zhang2025pomato}, and ZeroMSF~\cite{liang2025zero}, along with concurrent works Any4D~\cite{karhade2025any4d} and V-DPM~\cite{sucar2026vdpm}. We evaluate 3D tracking results in Tab.~\ref{tab:3dtracking}, following the approach described in Sec.~\ref{temporaltracking} to achieve dense tracking and perform sampling based on the UV coordinates of the query points. The results demonstrate that our method consistently outperforms most existing 3D tracking and pairwise scene flow approaches across multiple datasets, in both camera-centric and world-centric coordinate systems, highlighting its effectiveness and generalization.

\subsection{2D Tracking Estimation}
\textbf{Evaluation datasets.} Following prior works~\cite{karaev2024cotracker3,karaev2024cotracker,doersch2024bootstap}, we evaluate our method on three datasets: Kinetics~\cite{carreira2017quo}, RoboTAP~\cite{vecerik2024robotap}, and RGB-Stacking~\cite{lee2021beyond}. Kinetics comprises 1,144 YouTube videos from the Kinetics-700–2020 validation set, featuring complex camera motion and cluttered backgrounds, with an average of 26 tracks per video. RoboTAP consists of 265 real-world robotic manipulation videos, with an average duration of 272 frames per video. RGB-Stacking is a synthetic dataset of robotic videos, characterized by numerous texture-less regions that make tracking particularly challenging.

\noindent\noindent\textbf{Evaluation metrics.} We evaluate tracking performance using the TAP-Vid metrics~\cite{doersch2022tap}. Specifically, we report Occlusion Accuracy (OA), measuring occlusion prediction as a binary classification; $\delta_{\text{vis}}^{\text{avg}}$, the fraction of visible points tracked within 1, 2, 4, 8, and 16 pixels, averaged across thresholds; and Average Jaccard (AJ), which jointly assesses geometric and occlusion prediction accuracy.

\noindent\textbf{Comparison with existing methods.} We compare our method with representative 2D tracking approaches, including CoTracker3~\cite{karaev2024cotracker3}, LocoTrack~\cite{cho2024local}, and other state-of-the-art methods. In Tab.~\ref{tab:tab_tapvid_three}, we present the 2D tracking results from our 2D branch. Leveraging valuable geometric cues and joint training on multi-modal, multi-source data, our method achieves performance comparable to existing state-of-the-art 2D tracking approaches.
\begin{table}[t]
\centering
\setlength{\tabcolsep}{3pt} 
\footnotesize
\begin{adjustbox}{width=\textwidth}
\begin{tabular}{lcccccccccc}
\toprule
\multirow{2}{*}[-0.2em]{Method} 
& \multicolumn{3}{c}{Kinetics~\cite{carreira2017quo}} 
& \multicolumn{3}{c}{RoboTAP~\cite{vecerik2024robotap} } 
& \multicolumn{3}{c}{RGB-S~\cite{lee2021beyond}}  \\
\cmidrule(lr){2-4}
\cmidrule(lr){5-7}
\cmidrule(lr){8-10}
& AJ$\uparrow$ & $\delta_{\text{vis}}^{\text{avg}}$$\uparrow$ & OA$\uparrow$ 
& AJ$\uparrow$ & $\delta_{\text{vis}}^{\text{avg}}$$\uparrow$ & OA$\uparrow$
& AJ$\uparrow$ & $\delta_{\text{vis}}^{\text{avg}}$$\uparrow$ & OA$\uparrow$ \\
\midrule
PIPs++~\cite{zheng2023pointodyssey}  & / & 63.5 & / & / & 63.0 & / & / & 58.5 & / \\
TAPIR~\cite{doersch2023tapir}        & 49.6 & 64.2 & 85.0 & 59.6 & 73.4 & 87.0 & 55.5 & 69.7 & 88.0 \\
CoTracker~\cite{karaev2024cotracker} & 49.6 & 64.3 & 83.3 & 58.6 & 70.6 & 87.0 & 67.4 & 78.9 & 85.2 \\
TAPTR~\cite{li2024taptr}             & 49.0 & 64.4 & 85.2 & 60.1 & 75.3 & 86.9 & 60.8 & 76.2 & 87.0 \\
LocoTrack~\cite{cho2024local}        & 52.9 & 66.8 & 85.3 & 62.3 & 76.2 & 87.1 & 69.7 & 83.2 & 89.5 \\
BootsTAPIR~\cite{doersch2024bootstap}& 54.6 & 68.4 & 86.5 & 64.9 & \cellcolor{second}{80.1} & 86.3 & 70.8 & 83.0 & 89.9 \\
CoTracker3~\cite{karaev2024cotracker3}& \cellcolor{second}{55.8} & \cellcolor{second}{68.5} & \cellcolor{second}{88.3}  & \cellcolor{second}{66.4} & 78.8 & \cellcolor{second}{90.8} & \cellcolor{second}{71.7} & \cellcolor{second}{83.6} & \cellcolor{second}{91.1} \\
\midrule
\textbf{Ours}  &\cellcolor{best}{59.1}  & \cellcolor{best}{71.3} & \cellcolor{best}{90.6}& \cellcolor{best}{70.9} & \cellcolor{best}{81.8} & \cellcolor{best}{93.3} & \cellcolor{best}{78.2} & \cellcolor{best}{88.5} & \cellcolor{best}{92.3} \\
\bottomrule
\end{tabular}
\end{adjustbox}
\caption{
\textbf{2D tracking estimation.} We evaluate our model on three datasets: Kinetics~\cite{carreira2017quo}, RoboTAP~\cite{vecerik2024robotap}, and RGB-Stacking~\cite{lee2021beyond}.
}
\label{tab:tab_tapvid_three}
\vspace{-0.6em}
\end{table}

\begin{table*}[!ht]
\centering
\caption{
\textbf{Evaluation on point map estimation.} 
Results are aligned with the ground truth by optimizing a shared scale factor and shift across the entire video.
}
\vspace{-0.6em}
\label{tab:depth_pointmap}
\begin{adjustbox}{width=\textwidth}
\begin{tabular}{l c cc cc cc cc cc cc cc cc}
\toprule
\multirow{2}{*}{Method} & \multirow{2}{*}{Params} 
& \multicolumn{2}{c}{GMU Kitchen~\cite{georgakis2016multiview}} 
& \multicolumn{2}{c}{Monkaa~\cite{mayer2016large}} 
& \multicolumn{2}{c}{Sintel~\cite{butler2012naturalistic}} 
& \multicolumn{2}{c}{Scannet test~\cite{dai2017scannet}} 
& \multicolumn{2}{c}{Kubric-3D val~\cite{greff2022kubric}} 
& \multicolumn{2}{c}{KITTI~\cite{geiger2013vision}} 
& \multicolumn{2}{c}{Tum~\cite{sturm2012benchmark}} 
& \multicolumn{2}{c}{\textbf{Avg.}} \\
\cmidrule(lr){3-4} \cmidrule(lr){5-6} \cmidrule(lr){7-8} \cmidrule(lr){9-10} \cmidrule(lr){11-12} \cmidrule(lr){13-14} \cmidrule(lr){15-16} \cmidrule(lr){17-18}
& & Abs Rel $\downarrow$ & $\delta<1.25 \uparrow$ 
& Abs Rel $\downarrow$ & $\delta<1.25 \uparrow$ 
& Abs Rel $\downarrow$ & $\delta<1.25 \uparrow$ 
& Abs Rel $\downarrow$ & $\delta<1.25 \uparrow$ 
& Abs Rel $\downarrow$ & $\delta<1.25 \uparrow$ 
& Abs Rel $\downarrow$ & $\delta<1.25 \uparrow$ 
& Abs Rel $\downarrow$ & $\delta<1.25 \uparrow$  
& Abs Rel $\downarrow$ & $\delta<1.25 \uparrow$ \\
\midrule
MoGe~\cite{wang2025moge1} & 314M & 0.1728 & 0.6725 & 0.2069 & 0.6317 & 0.2181 & 0.6615 & 0.1194 & 0.8447 & 0.0852 & 0.9294 & 0.0801 & 0.9374 & 0.1563 & 0.7663 & 0.1484 & 0.7776 \\
VGGT~\cite{wang2025vggt} & 1.26B & 0.2530 & 0.4454 & 0.2009 & 0.6678 & 0.2004 & 0.7303 & 0.0763 & 0.9242 & \cellcolor{second}{0.0316} & 0.9733 & 0.1245 & 0.8517 & 0.1278 & 0.8236 & 0.1449 & 0.7738 \\
MoGe-2~\cite{wang2025moge} & 331M & 0.0654 & \cellcolor{best}{0.9391} & 0.1904 & 0.6797 & 0.2058 & 0.6903 & 0.0626 & 0.9673 & 0.1184 & 0.8592 & 0.0776 & 0.9627 & 0.1156 & 0.8694 & 0.1194 & 0.8525 \\
MapAnything~\cite{keetha2025mapanything} & 563M & 0.1090 & 0.9279 & 0.2012 & 0.7379 & 0.2059 & 0.7141 & 0.0463 & 0.9833 & 0.0684 & 0.9475 & 0.1016 & 0.9144 & 0.1132 & 0.9458 & 0.1208 & 0.8816 \\
Pi3~\cite{wang2025pi} & 1.29B & \cellcolor{second}{0.0458} & 0.9338 & \cellcolor{best}{0.0774} & \cellcolor{best}{0.9256} & \cellcolor{second}{0.1489} & 0.7899 & 0.0750 & 0.9532 & 0.0337 & \cellcolor{second}{0.9817} & 0.0866 & 0.8877 & \cellcolor{best}{0.0493} & \cellcolor{best}{0.9680} & \cellcolor{second}{0.0738} & 0.9200 \\
DA3~\cite{lin2025depth} & 1.36B & 0.0551 & \cellcolor{second}{0.9343} & 0.1131 & 0.8761 & 0.1575 & \cellcolor{second}{0.8064} & \cellcolor{best}{0.0222} & \cellcolor{best}{0.9917} & 0.0431 & 0.9599 & \cellcolor{second}{0.0404} & \cellcolor{second}{0.9861} & 0.0886 & 0.9515 & 0.0743 & \cellcolor{second}{0.9294} \\
\midrule
\textbf{Ours} & 1.38B & \cellcolor{best}{0.0431} & 0.9341 & \cellcolor{second}{0.0853} & \cellcolor{second}{0.9123} & \cellcolor{best}{0.1261} & \cellcolor{best}{0.8291} & \cellcolor{second}{0.0288} & \cellcolor{second}{0.9872} & \cellcolor{best}{0.0191} & \cellcolor{best}{0.9939} & \cellcolor{best}{0.0268} & \cellcolor{best}{0.9877} & \cellcolor{second}{0.0572} & \cellcolor{second}{0.9638} & \cellcolor{best}{0.0552} & \cellcolor{best}{0.9440} \\
\bottomrule
\end{tabular}
\end{adjustbox}
\vspace{-0.6em}
\end{table*}

\subsection{Point Map Estimation}

\textbf{Evaluation datasets.} We evaluate our model on seven datasets, covering both real-world and synthetic scenarios. For indoor real-world datasets, we use GMU Kitchen~\cite{georgakis2016multiview}, ScanNet test set~\cite{dai2017scannet}, and TUM~\cite{sturm2012benchmark}. For outdoor real-world data, we use KITTI~\cite{geiger2013vision}. For synthetic datasets, we evaluate on Monkaa~\cite{mayer2016large}, Sintel~\cite{butler2012naturalistic}, and Kubric-3D validation set~\cite{greff2022kubric}.

\noindent\textbf{Evaluation metrics.} Following prior works~\cite{hu2024depthcrafter,zhang2024monst3r,lu2024align3r}, we first align the estimated point/depth maps with the ground truth using a shared scale and shift before computing the metrics. This alignment procedure follows the approach described in MoGe~\cite{wang2025moge1}. We primarily report two metrics: the absolute relative error (Abs Rel) and the percentage of inlier points within a threshold $\delta < 1.25$.

\noindent\textbf{Comparison with existing methods.}
We compare our method with several representative approaches, including MoGe~\cite{wang2025moge1}, VGGT~\cite{wang2025vggt}, MapAnything~\cite{keetha2025mapanything}, Pi3~\cite{wang2025pi}, and DA3~\cite{lin2025depth}. Tab.~\ref{tab:depth_pointmap} presents a quantitative comparison against these methods. As shown, our method achieves highly competitive geometry estimation performance due to our tailored training strategies, providing a solid foundation for 3D flow estimation.

\subsection{Camera Pose Estimation}
\textbf{Evaluation datasets.} 
We evaluate camera pose estimation performance on two widely used benchmarks: Sintel~\cite{butler2012naturalistic} and Bonn~\cite{palazzolo2019refusion}. 
For the Sintel dataset, following prior works~\cite{chen2024leap,zhao2022particlesfm,zhang2024monst3r}, 
we exclude sequences that are static or contain only straight motions, resulting in 14 dynamic sequences for evaluation. 

\noindent\textbf{Evaluation metrics.} 
Consistent with existing works~\cite{chen2024leap,zhao2022particlesfm,teed2024deep,zhang2024monst3r}, 
we report three standard metrics: 
ATE $\downarrow$ (Absolute Translation Error), 
RTE $\downarrow$ (Relative Translation Error), 
and RRE $\downarrow$ (Relative Rotation Error). 

\noindent\textbf{Comparison with existing methods.} We compare our approach with several joint depth and pose estimation methods, including Align3R~\cite{lu2024align3r}, CUT3R~\cite{wang2025continuous}, VGGT~\cite{wang2025vggt}, MapAnything~\cite{keetha2025mapanything}, Pi3~\cite{wang2025pi}, and DA3~\cite{lin2025depth}, as well as joint depth, pose, and motion estimation methods, such as POMATO~\cite{zhang2025pomato} and STV2~\cite{xiao2025spatialtrackerv2}.
Tab.~\ref{pose_estimation} reports the camera pose estimation results.
The results demonstrate the effectiveness of our method, particularly among approaches that jointly estimate depth, pose, and motion.
\begin{table}[!ht]
  \begin{center}
  \vspace{-0.6em}
    \footnotesize
    \caption{\textbf{Camera pose estimation.} We evaluate our model on two datasets: Sintel~\cite{butler2012naturalistic} and Bonn~\cite{palazzolo2019refusion}. $\ddagger$ indicates using ground-truth camera intrinsics as input.}
    \label{pose_estimation}
    \begin{adjustbox}{width=\textwidth}
    \begin{tabular}{lccc|ccc}
      \toprule
 \multirow{2}{*}{Method} 
      & \multicolumn{3}{c|}{Sintel~\cite{butler2012naturalistic}} 
      & \multicolumn{3}{c}{Bonn~\cite{palazzolo2019refusion}}\\
       & ATE$\downarrow$ & RTE$\downarrow$ & RRE$\downarrow$
        & ATE$\downarrow$ & RTE$\downarrow$ & RRE$\downarrow$ \\
      \midrule
      
      Align3R~\cite{lu2024align3r} & 0.128 & \cellcolor{best}{0.042} & 0.432 & 0.023 & \cellcolor{best}{0.007} & 0.620  \\
      CUT3R~\cite{wang2025continuous} & 0.217 & 0.070 & 0.636 & 0.035 & 0.014 & 1.212 \\

      VGGT~\cite{wang2025vggt} & 0.167 & 0.062 & 0.490 & 0.051 & 0.011 & 1.038\\
      MapAnything~\cite{keetha2025mapanything} &0.227 & 0.111&2.047&0.026 &0.014 &0.668 \\
      Pi3~\cite{wang2025pi} &   \cellcolor{best}{0.088} & \cellcolor{second}{0.043} & \cellcolor{best}{0.299} &0.012&0.011&\cellcolor{second}{0.612}\\ 
DA3~\cite{lin2025depth}          
& 0.124 & 0.061 & 0.331&\cellcolor{second}{0.010}&0.011&0.638 \\
\midrule
POMATO~\cite{team2025aether} & 0.209 & 0.064 & 0.694 & 0.041 & 0.017 & 0.832  \\
STV2~\cite{xiao2025spatialtrackerv2} 
& 0.133 & 0.057 & 0.641&0.019&0.015&0.701 \\
\textbf{Ours} 
& \cellcolor{second}{0.119} & 0.054 & \cellcolor{second}{0.309}&\cellcolor{best}{0.009}&\cellcolor{second}{0.009}&\cellcolor{best}{0.604}  \\

      \bottomrule
    \end{tabular}
     \end{adjustbox}
     \vspace{-2em}
  \end{center}
\end{table}

\subsection{Ablation Study}

\noindent\textbf{Effect of different 3D backbone ViT models.}
Tab.~\ref{tab:diff_found} reports the average results obtained using different 3D backbone ViT models as the global geometry encoder initialization.
The results demonstrate the flexibility of our framework and show that it remains effective across different backbone choices.

\begin{table}[!ht]
\centering
% \vspace{-0.6em}
\caption{\textbf{Effect of different 3D foundation models.}}
% \vspace{-0.6em}
\label{tab:diff_found}
\begin{adjustbox}{width=1\textwidth}
\begin{tabular}{
    lcccccccc
}
\toprule
& \multicolumn{2}{c}{Point Map} 
& \multicolumn{3}{c}{Scene Flow} 
& \multicolumn{2}{c}{3D Tracking} \\
Method 
& Abs Rel $\downarrow$ 
& $\delta<1.25 \uparrow$ 
& EPE3D $\downarrow$ 
& Abs Rel $\downarrow$ 
& $\delta<1.25 \uparrow$ 
& L-16 $\uparrow$ 
& L-50 $\uparrow$ \\ 
\midrule
Ours (MoGe~\cite{wang2025moge1}) 
& 0.0973 & 0.8921 
& 0.3180 & 0.0680 & 0.9479  
& 0.5447 & 0.5135 \\
Ours (Pi3~\cite{wang2025pi}) 
&\cellcolor{best}{0.0492}&\cellcolor{best}{0.9537} 
& \cellcolor{second}{0.2569} & \cellcolor{second}{0.0548} & \cellcolor{best}{0.9645}  
& \cellcolor{best}{0.5734} &\cellcolor{second}{0.5424} \\
Ours (DA3~\cite{lin2025depth}) 
& \cellcolor{second}{0.0552} & \cellcolor{second}{0.9440}
& \cellcolor{best}{0.2056} & \cellcolor{best}{0.0474} & \cellcolor{second}{0.9607} 
& \cellcolor{second}{0.5712} & \cellcolor{best}{0.5469} \\
\bottomrule
\end{tabular}
\end{adjustbox}
\vspace{-0.6em}
\end{table}

\noindent \textbf{Ablation study on the scene flow decoder.}
Tab.~\ref{table:ablation} validates our key design choices. 
\textbf{(1) 2D Supervision:} Training solely on 3D datasets (``w/o 2D Supervision'') causes a severe performance collapse (EPE3D 0.6511), underscoring its necessity for guiding 3D flow. 
\textbf{(3) Target Lifting:} Removing lifted target samples (``w/o Target lifting'') prevents the network from dynamically querying the target space via 2D displacements. The resulting performance drop confirms that mapping 2D matches to 3D residuals is essential for spatial refinement. 
\textbf{(4) Iterative Updates:} Replacing the iterative refinement mechanism with a direct, single-step regression (``w/o iterations'') leads to a performance decline. This highlights that iteratively updating the flow field is crucial for achieving highly precise results. 
\textbf{(5) Auxiliary Priors:} Ablating the auxiliary priors (``w/o $\mathbf{C}^{(t)}_{3d,i,j}$ \& $\mathcal{M}_{3d}^{(t)}$'') decreases accuracy, verifying their indispensability for global spatial alignment and kinematic smoothness across iterations.
\textbf{(4) Hybrid Formulation (introduced in Sec~\ref{dense_recovery}):} Our full model (EPE3D 0.2056) significantly outperforms both heuristically lifting 2D flow with depth (``2D flow + $d$'', 0.8210) and purely regressing 3D flow (``Pure 3D flow'', 0.2854). This proves that explicitly intertwining 2D flow with Z-axis displacement is superior to relying on isolated domains.

\begin{table}[t]
  \centering
  \footnotesize
  \setlength\tabcolsep{4pt}
  \renewcommand{\arraystretch}{1.1}
  \caption{\textbf{Ablation study on the scene flow decoder.} The results are averaged over all datasets. \colorbox{best}{Best} and \colorbox{second}{second best} results are highlighted.\vspace{-1em}}
  \label{table:ablation}
  \begin{adjustbox}{width=0.8\linewidth}
  \begin{tabular}{lccc}
    \toprule
    Setting & Abs Rel $\downarrow$ & $\delta < 1.25 \uparrow$ & EPE3D $\downarrow$ \\
    \midrule
    \multicolumn{4}{l}{\textit{Supervision}} \\
    w/o 2D Supervision        & 0.1021 & 0.9199 & 0.6511 \\

    \midrule
    \multicolumn{4}{l}{\textit{Decoder Components}} \\
    w/o Target lifting        & 0.0534 & 0.9543 & 0.3017 \\
    w/o iterations            & 0.0506 & 0.9559 & 0.2356 \\
    w/o $\mathbf{C}^{(t)}_{3d,i,j}$ \& $\mathcal{M}_{3d}^{(t)}$ & \cellcolor{second}{0.0482} & \cellcolor{second}{0.9591} & \cellcolor{second}{0.2201} \\
    
    \midrule
    \multicolumn{4}{l}{\textit{Hybrid Formulation}} \\
    2D flow + $d$             & 0.1304 & 0.8911 & 0.8210 \\
    Pure 3D flow              & 0.0552 & 0.9570 & 0.2815 \\
    
    \midrule
    \textbf{Full (Ours)}      & \cellcolor{best}{0.0474} & \cellcolor{best}{0.9607} & \cellcolor{best}{0.2056} \\
    \bottomrule
  \end{tabular}
  \end{adjustbox}
  \vspace{-0.6em}
\end{table}

\noindent\textbf{Efficiency of scene flow decoder.}
Tab.~\ref{tab:efficiency1} compares the inference time (Time), memory consumption (Mem.), and parameter count (Parm.) on 16-frame sequences from the ADT~\cite{pan2023aria} dataset. 
First, pairwise scene flow approaches, such as POMATO~\cite{zhang2025pomato} and ZeroMSF~\cite{liang2025zero}, natively support dense tracking by computing flow independently for every pixel. However, unlike our efficient sparse-to-dense formulation, their per-pixel prediction approach incurs significantly higher computational cost, resulting in increased latency without providing a clear advantage in memory efficiency. 
On the other hand, STV2~\cite{xiao2025spatialtrackerv2} relies on a traditional 3D spatial correlation mechanism for iterative trajectory estimation. Its high computational complexity and large parameter count impose a severe bottleneck. Consequently, STV2 is restricted to sparse tracking and suffers from Out-of-Memory (OOM) errors when extended to dense tracking scenarios. 
To explicitly validate that this bottleneck lies in the traditional 3D computation, we replace our proposed 2D-to-3D correlation module with the aforementioned traditional 3D mechanism (\textbf{Ours w/o 2D-to-3D}). As shown in the gray rows of Tab.~\ref{tab:efficiency1}, this variant similarly crashes with an OOM error under dense tracking settings. 
Overall, by bypassing the prohibitive cost of explicit 3D correlations, our method achieves superior efficiency in runtime, memory footprint, and model size, all while natively supporting dense tracking and maintaining state-of-the-art performance.

\begin{table}[!ht]
\centering
\caption{\textbf{Efficiency comparison on 16-frame ADT~\cite{pan2023aria} sequences.} Gray rows indicate Out-of-Memory (OOM) failures under dense tracking settings.}
\label{tab:efficiency1}
\begin{adjustbox}{width=0.8\textwidth}
\begin{tabular}{lccc}
\toprule
Method & Time (s) $\downarrow$ & Mem. (GB) $\downarrow$ & Parm. (M) $\downarrow$ \\
\midrule
POMATO~\cite{zhang2025pomato} (Dense) &  \cellcolor{second}{4.8} & 16 & 133.64  \\
ZeroMSF~\cite{liang2025zero} (Dense) & 8.2 & \cellcolor{best}{10} & 153.84 \\
STV2~\cite{xiao2025spatialtrackerv2} (Sparse) & 5.8 & 19 & 65.99 \\
\midrule
\textcolor{gray}{\textbf{STV2~\cite{xiao2025spatialtrackerv2}} (Dense)} & \textcolor{gray}{OOM} & \textcolor{gray}{OOM} &\textcolor{gray}{65.99} \\
\textcolor{gray}{\textbf{Ours w/o 2D-to-3D} (Dense)} & \textcolor{gray}{OOM} & \textcolor{gray}{OOM} &\textcolor{gray}{56.90} \\
\midrule
\textbf{Ours} (Dense) & \cellcolor{best}{3.4} &  \cellcolor{second}{14} & \cellcolor{best}{26.06} \\
\bottomrule
\end{tabular}
\end{adjustbox}
\vspace{-0.6em}
\end{table}

 \begin{figure}[!ht]
    % \vspace{-0.4em}
    \begin{center}
        \includegraphics[width=0.96\linewidth]{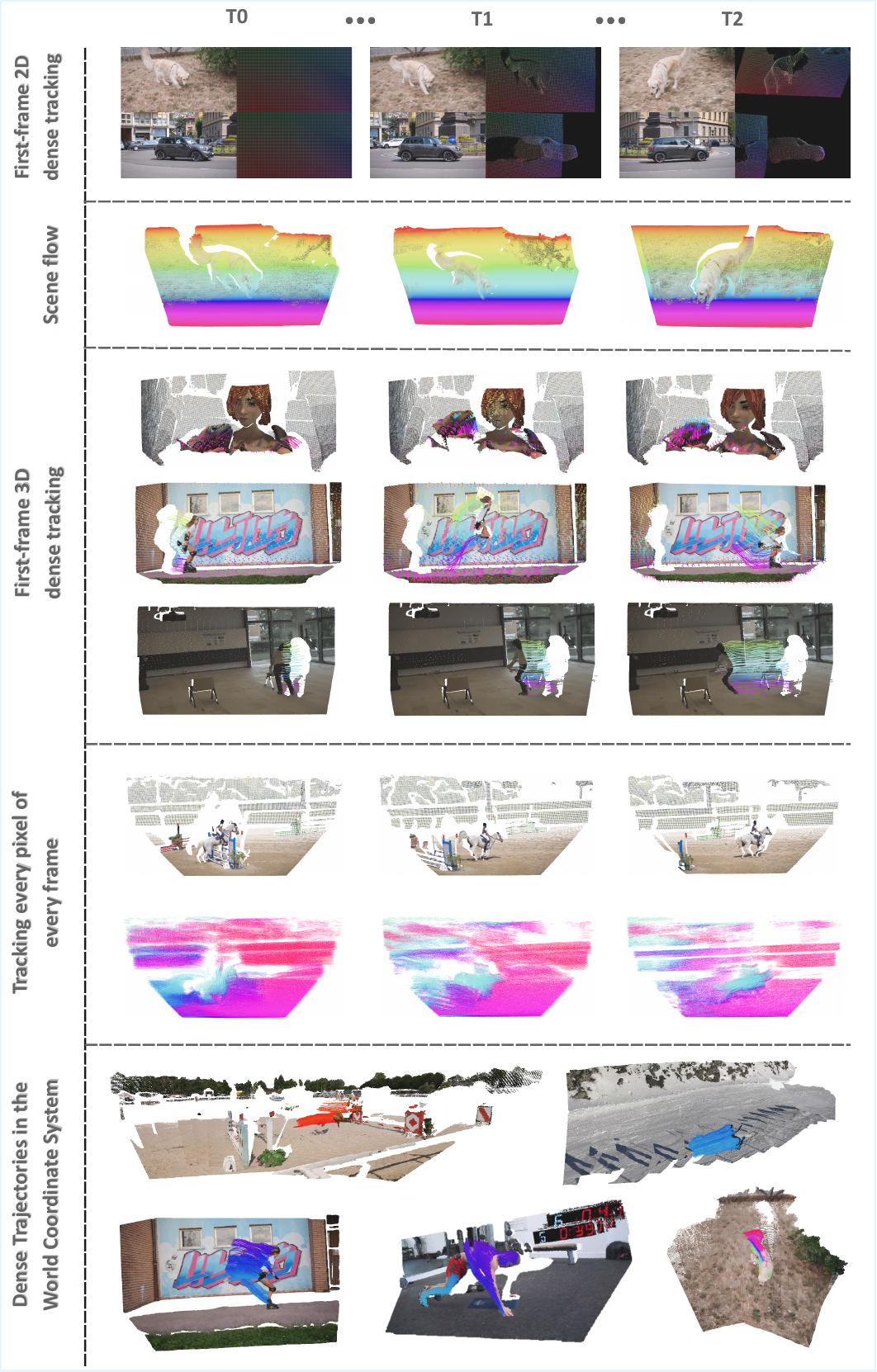}
\caption{\textbf{Qualitative results on diverse in-the-wild videos.} }
        \label{fig:vis}
    \end{center}
    \vspace{-1.5em}
\end{figure}

\subsection{Qualitative Visualization}
To demonstrate our method's effectiveness, Fig.~\ref{fig:vis} visualizes the tracking and flow results across five aspects. The first four illustrate \textbf{camera-centric} perspectives:
(1) \textbf{1st-frame 2D tracking}, where consistent colors denote temporal pixel correspondences.
(2) \textbf{Scene flow}, contrasting the original RGB point cloud with the flow-transformed geometry (rainbow).
(3) \textbf{1st-frame 3D tracking}, featuring uniformly downsampled camera-centric trajectories for rendering efficiency.
(4) \textbf{All-frame 3D tracking}, capturing dense camera-centric trajectories of both existing and newly emerging objects.
Distinct from the above, (5) \textbf{World-centric tracking} maps these trajectories to a global coordinate system. This effectively decouples camera ego-motion from object dynamics, yielding spatially stable backgrounds and physically coherent absolute motions for dynamic entities.
For dynamic visualizations, please refer to the supplementary video.

\section{Conclusion}

In this paper, we present \textbf{Track4World}, a feedforward foundation model designed for efficient, dense 3D tracking within a world-centric coordinate framework. Leveraging a global 3D scene representation parameterized by a VGGT-style Vision Transformer, our method introduces a novel 3D correlation mechanism to simultaneously estimate dense 2D and 3D scene flow between arbitrary pairs of frames. By tightly coupling this estimated scene flow with the reconstructed 3D geometry, Track4World facilitates highly efficient, pixel-wise 3D trajectory extraction across the entire video. Extensive evaluations across multiple benchmarks demonstrate the exceptional robustness, scalability, and generalization capabilities of our approach, marking a significant step toward holistic 4D reconstruction from in-the-wild monocular videos.
\clearpage
\setcounter{page}{1}
\maketitlesupplementary
\appendix
\renewcommand{\figurename}{Fig.}
\renewcommand{\thefigure}{S\arabic{figure}}
\setcounter{figure}{0}

\renewcommand{\tablename}{Table}
\renewcommand{\thetable}{S\arabic{table}}
\setcounter{table}{0}
\section{Overview}
In this supplementary material, we provide comprehensive details regarding the implementation, training, and evaluation of our proposed method. The document is organized as follows:

\begin{itemize}
    \item \textbf{Sec.~\ref{sec:impl_details}} provides the specific implementation details, including training hyperparameters, dataset configurations, and optimization strategies.
    \item \textbf{Sec.~\ref{sec:arch_details}} elaborates on the detailed network architecture of the flow estimation module.
    \item \textbf{Sec.~\ref{sec:losses}} details the unified objective functions designed for joint flow estimation and point tracking.
    \item \textbf{Sec.~\ref{sec:pose_est}} describes the optimization procedure and mathematical formulation used for camera pose estimation.
    \item \textbf{Sec.~\ref{sec:metrics}} defines the evaluation metrics used across geometry, flow, tracking, and camera pose tasks to ensure clarity and reproducibility.
    \item \textbf{Sec.~\ref{sec:more_ablation}} presents additional ablation studies to further validate our design choices.
    \item \textbf{Sec.~\ref{sec:more_vis}} showcases extensive qualitative results visualizations on challenging in-the-wild sequences.
    \item \textbf{Sec.~\ref{sec:limitations}} discusses the constraints of the current approach and outlines promising directions for future research.
\end{itemize}
\section{Implementation Details}
\label{sec:impl_details}
Our model is trained in two stages. In the first stage, we focus on geometry estimation, using a diverse set of datasets that provide high-quality depth and pose supervision, including Kubric-3D~\cite{greff2022kubric}, GTA-SfM~\cite{wang2020flow}, V-KITTI~\cite{gaidon2016virtual}, ARKitScenes~\cite{baruch2021arkitscenes}, BlinkVision~\cite{li2024blinkvision}, DynamicStereo~\cite{karaev2023dynamicstereo}, TartanAir~\cite{wang2020tartanair}, MVS-Synth~\cite{huang2018deepmvs}, ScanNetv2~\cite{dai2017scannet}, ScanNet++~\cite{yeshwanth2023scannet++}, Spring~\cite{mehl2023spring}, OmniWorld~\cite{zhou2025omniworld}, PointOdyssey~\cite{zheng2023pointodyssey}, Hypersim~\cite{roberts2021hypersim}, and IRS~\cite{wang2019irs}.
In the second stage, we freeze the geometry estimation module and train the motion estimation module. The training data covers a wide range of motion patterns, including optical flow datasets (AutoFlow~\cite{sun2021autoflow}, FlyingChairs~\cite{dosovitskiy2015flyingchairs}, OmniWorld~\cite{zhou2025omniworld}, HD1K~\cite{kondermann2016hd1k}, Spring~\cite{mehl2023spring}, TartanAir~\cite{wang2020tartanair}, VIPER~\cite{richter2017playing}), 2D point tracking datasets (Kubric~\cite{greff2022kubric}, Monkaa~\cite{mayer2016large}, Driving~\cite{mayer2016large}), scene flow datasets (Kubric-3D~\cite{greff2022kubric}, V-KITTI~\cite{gaidon2016virtual}), and 3D point tracking datasets (DynamicStereo~\cite{karaev2023dynamicstereo}, PointOdyssey~\cite{zheng2023pointodyssey}).

We train our model on eight 40GB GPUs. 
In the first stage, we adopt the AdamW~\cite{loshchilov2017decoupled} optimizer with a StepLR scheduler to train the geometry estimation backbone, 
using an initial learning rate of $1\times10^{-4}$. 
This stage runs for 100{,}000 steps and takes approximately one week. 
In the second stage, we train the motion estimation module using the AdamW optimizer with a OneCycleLR scheduler, 
whose peak learning rate is set to $1\times10^{-4}$. 
This stage is trained for another 100{,}000 steps and takes about five days.

\section{Finetuning Geometry Encoder}

We employ a backbone-agnostic encoder to process video frames $\{\mathbf{I}_i\}$ into a global scene representation. Leveraging state-of-the-art models~\cite{wang2025pi,lin2025depth,wang2025moge1} as initialization, we fine-tune specific layers to enhance temporal consistency: adding global attention layers and camera pose tokens for monocular models like MoGe~\cite{wang2025moge1}, or updating intermediate layers for 3D reconstruction models like Pi3~\cite{wang2025pi} and DA3~\cite{lin2025depth}. As shown in Tab.~\ref{tab:diff_found}, we demonstrate the effectiveness of our proposed framework across different geometry backbones.

To address the
inherent scale and focal-length ambiguity while ensuring temporal coherence, we compute a video-level reconstruction loss. We estimate a global optimal scale $s^*$ and translation $t^*$ to align the predicted point clouds $\mathbf{P}_i$ with the ground truth $\hat{\mathbf{P}}_i$:
\begin{equation}
\ell_{recons} = \frac{1}{|T|} \sum_{i \in T} \frac{1}{|\mathcal{M}_i|}
\sum_{j \in \mathcal{M}_i} \frac{1}{z_{i,j}} \left\| s^* \mathbf{p}_{i,j} + t^* - \hat{\mathbf{p}}_{i,j} \right\|_1,
\end{equation}
where $z_{i,j}$ denotes the ground-truth depth used for re-weighting. Furthermore, we enforce global geometric consistency via an affine-invariant pairwise pose loss:
\begin{equation}
\begin{split}
\ell_{pose} = \frac{1}{N(N-1)} \sum_{i \neq j} \Big( & \ell_{geo}(\mathbf{R}_{ij}, \hat{\mathbf{R}}_{ij}) \\
& + \lambda_{trans} \ell_{huber}( \tilde{\mathbf{t}}_{ij}, \hat{\mathbf{t}}_{ij} ) \Big),
\end{split}
\end{equation}
where predicted translations are aligned as $\tilde{\mathbf{t}}_{ij} = s^* \mathbf{t}_{ij} + t^* - \hat{\mathbf{R}}_{ij} t^*$.
To stabilize convergence, we impose a regularization $\ell_{reg} = \max(0, L_{mean} - \tau)$ on the mean point magnitude $L_{mean}$.
For geometric fidelity, we further introduce a normal consistency loss $\ell_n$ and a multi-scale local geometry loss $\ell_{local}$~\cite{wang2025moge1}. The overall training objective is:
\begin{equation}
\ell_{total} = \ell_{recons} + \lambda_{pose} \ell_{pose} + \lambda_{reg} \ell_{reg} + \lambda_n \ell_n + \lambda_{local} \ell_{local}.
\end{equation}

\section{Model Architecture Details}
\label{sec:arch_details}
\paragraph{Recurrent update operator $\mathrm{GRU}$.}
The update operator $\mathrm{GRU}$ refines the motion representations $\mathbf{M}_{2d}^{(t)}$ and $\mathbf{V}^{(t)}$ by integrating correlation cues, recurrent features, semantic context, and temporal motion variation.
At iteration $(t)$, the operator takes as inputs:
the 4D correlation volumes $\hat{\mathbf{C}}_{i,j}(\mathbf{M}_{2d}^{(t)})$ and $\tilde{\mathbf{C}}_{i,j}(\mathbf{M}_{2d}^{(t)})$,
the motion-change gradient $\nabla(\mathbf{M}_{2d}^{(t)} - \mathbf{M}_{2d}^{(t-1)})$,
the current recurrent feature $\hat{\mathbf{F}}^{(t)}_i$,
the semantic context feature $\tilde{\mathbf{F}}_i$,
and the current visibility confidence map $\mathbf{V}^{(t)}$.
First, motion encoders $\phi_{\text{mot}}$ extracts correlation-based motion cues from the correlation volumes:
\begin{align}
\mathbf{F}^{\text{corr}}_{i,j} 
&= \phi_{\text{mot}1}\!\left(
    \hat{\mathbf{C}}_{i,j},
    \nabla\!\left(\mathbf{M}_{2d}^{(t)} - \mathbf{M}_{2d}^{(t-1)}\right)
\right), \\
\mathbf{F}^{\text{ctx}}_{i,j} 
&= \phi_{\text{mot}2}\!\left(
    \tilde{\mathbf{C}}_{i,j},
    \nabla\!\left(\mathbf{M}_{2d}^{(t)} - \mathbf{M}_{2d}^{(t-1)}\right)
\right).
\end{align}
Simultaneously, the recurrent feature and the context feature are projected into a shared latent space via respective functions $\psi_f$ and $\psi_c$:
\begin{equation}
\hat{\mathbf{F}}^{(t),\text{corr}}_i = \psi_f(\hat{\mathbf{F}}^{(t)}_i), \qquad
\tilde{\mathbf{F}}^{\text{ctx}}_i = \psi_c(\tilde{\mathbf{F}}_i).
\end{equation}
These processed features are then concatenated with the unprocessed inputs, namely the motion-change gradient and the visibility map. This aggregated tensor is fused via a $1{\times}1$ convolution $\psi$ to form the initial hidden state $\mathbf{H}_0$:
\begin{equation}
\mathbf{H}_0 = \psi\Big(
[\hat{\mathbf{F}}^{(t),\text{corr}}_i, \tilde{\mathbf{F}}^{\text{ctx}}_i, \mathbf{F}^{\text{corr}}_{i,j}, \mathbf{F}^{\text{ctx}}_{i,j}, \mathbf{V}^{(t)}]
\Big).
\end{equation}

This initial state $\mathbf{H}_0$ is then refined through multi
\textit{temporal aggregator} (TA) layers to do attention along the temporal dimension, as introduced in Sec.~3.4:
\begin{equation}
\mathbf{H}_{k+1} = \text{TA}_k(\mathbf{H}_k),
\qquad k = 0,\dots,K{-}1.
\end{equation}
The resulting feature $\mathbf{H}_K$ is passed through a final $1{\times}1$ projection $\psi_{\text{final}}$ to produce the updated recurrent feature $\hat{\mathbf{F}}^{(t+1)}_i$ for the \emph{next} iteration ($t+1$):
\begin{equation}
\hat{\mathbf{F}}^{(t+1)}_i = \psi_{\text{final}}(\mathbf{H}_K).
\end{equation}
Finally, two lightweight MLP heads use this new hidden feature $\hat{\mathbf{F}}^{(t+1)}_i$ to predict the \emph{updates} ($\Delta\mathbf{M}_{2d}^{(t)}$ and $\Delta\mathbf{V}^{(t)}$) for the 2D motion and visibility fields:
\begin{equation}
\Delta\mathbf{M}_{2d}^{(t)} = \text{MLP}_{mot}(\hat{\mathbf{F}}^{(t+1)}_i), \quad
\mathbf{M}_{2d}^{(t+1)} = \mathbf{M}_{2d}^{(t)} + \Delta\mathbf{M}_{2d}^{(t)},
\end{equation}
\begin{equation}
\Delta\mathbf{V}^{(t)} = \text{MLP}_{vis}(\hat{\mathbf{F}}^{(t+1)}_i), \quad
\mathbf{V}^{(t+1)} = \mathbf{V}^{(t)} + \Delta\mathbf{V}^{(t)}.
\end{equation}

Overall, $\mathrm{GRU}$ functions as a recurrent refinement operator that jointly
aggregates correlation cues, recurrent and semantic features, temporal motion
differences, and visibility confidence. This enables robust and consistent iterative
updates of the dense motion and visibility fields.

\paragraph{3D flow head $\mathcal{H}_{3d}$.}
The 3D flow head $\mathcal{H}_{3d}$ aggregates all prepared features to predict the 3D flow update $\Delta \mathbf{M}_{3d}^{(t)}$.
Formally, given the lifted source features $\hat{\mathbf{F}}^{(t+1)}_i$, $\mathbf{F}_i$, the lifted target point coordinates $\mathbf{p}_j^{(t)}$, $\mathbf{p}_j^{(t+1)}$ and features $\mathbf{F}_j^{(t)}, \mathbf{F}_j^{(t+1)}$, the 3D-to-2D point correlation $\mathbf{C}_{3d,i,j}^{(t)}$, and the 3D flow prior $\mathcal{M}_{3d}^{(t)}$, the head operates as follows:
\begin{equation}
\begin{aligned}
\mathbf{F}'_i &= \phi_f^{3d}(\hat{\mathbf{F}}^{(t+1)}_i), &
\mathbf{F}'^{3d}_i &= \phi_s^{3d}(\mathbf{F}_i), \\
\mathbf{F}'^{(t)}_j &= \phi_p^{3d}(\mathbf{F}_j^{(t)}), &
\mathbf{F}'^{(t+1)}_j &= \phi_p^{3d}(\mathbf{F}_j^{(t+1)}), \\
\mathbf{M}^{3d}_{\text{corr}} &= \phi_{\text{mot}}^{3d}(\mathbf{C}_{3d,i,j}^{(t)}, \mathcal{M}_{3d}^{(t)}),
\end{aligned}
\end{equation}
where $\phi_f^{3d}$, $\phi_s^{3d}$, $\phi_p^{3d}$, and $\phi_{\text{mot}}^{3d}$ denote dedicated $1{\times}1$ convolutional projections or motion encoders. These features are concatenated and fused via a $1{\times}1$ convolution to form the initial hidden state:
\begin{equation}
\mathbf{H}_0^{3d} = \psi^{3d}\Big(
[\mathbf{F}'_i, \mathbf{F}'^{3d}_i, \mathbf{F}'^{(t)}_j, \mathbf{F}'^{(t+1)}_j, \mathbf{M}^{3d}_{\text{corr}}]
\Big).
\end{equation}
Temporal refinement is performed using a \textit{temporal aggregator} applied along the time dimension:
\begin{equation}
\mathbf{H}_{k+1}^{3d} = \text{TA}^{3d}(\mathbf{H}_k^{3d}), \quad k=0,\dots,K{-}1.
\end{equation}
After a final $1{\times}1$ projection, the refined hidden feature is concatenated with the 3D flow prior $\mathbf{p}_j^{(t+1)}-\mathbf{p}_j^{(t)}$:
\begin{equation}
\mathbf{F}^{3d}_{\text{out}} = [\psi_{\text{final}}^{3d}(\mathbf{H}_K^{3d}), \mathbf{p}_j^{(t+1)}-\mathbf{p}_j^{(t)}].
\end{equation}
Finally, a lightweight prediction head $\text{MLP}_{3d}$ composed of several convolutional layers outputs the 3D flow update:
\begin{equation}
\Delta \mathbf{M}_{3d}^{(t)} = \text{MLP}_{3d}(\mathbf{F}^{3d}_{\text{out}}), \quad\mathbf{M}_{3d}^{(t+1)} = \mathbf{M}_{3d}^{(t)} + \Delta\mathbf{M}_{3d}^{(t)}.
\end{equation}

Overall, $\mathcal{H}_{3d}$ serves as a 3D motion refinement module that jointly leverages 2D motion features, 3D source information, point correlations, temporal dependencies, and prior flow to produce accurate and temporally consistent 3D flow updates.
\section{Losses for Flow and Tracking Estimation}
\label{sec:losses}
We observe that both pairwise scene flow and long-term point tracking datasets share a similar data structure, characterized by point-wise correspondences across temporal sequences. Our arbitrary-pair querying mechanism fundamentally supports variable temporal intervals and flexible numbers of frames. Leveraging this structural similarity, we adopt a unified training objective for both tasks, ensuring consistent supervision regardless of the dataset source.
\subsection{Losses for 2D Branch}
For the 2D branch, we align our supervision strategy with state-of-the-art point tracking methods \cite{karaev2024cotracker, karaev2024cotracker3, harley2025alltracker}. The overall objective function, $\ell_{\text{total}}$, is computed over a sequence of $N_{iter}$ iteratively refined predictions, utilizing an exponential weighting factor $\gamma$ (typically set to 0.8) to emphasize the quality of later updates.
The total loss is a weighted sum of the trajectory loss ($\ell_{\text{traj}}$), the visibility loss ($\ell_{\text{vis}}$), and the confidence loss ($\ell_{\text{conf}}$), aggregated across all iterative steps $t$:

\begin{equation}
\ell_{\text{total}} = \sum_{t=1}^{N_{iter}} \gamma^{N_{iter}-t} \left( \ell_{\text{traj}}^{(t)} + \lambda_{\text{vis}} \ell_{\text{vis}}^{(t)} + \lambda_{\text{conf}} 
\ell_{\text{conf}}^{(t)} \right)
\end{equation}
\noindent \textit{Note: The weights $\lambda_{\text{vis}}$ and $\lambda_{\text{conf}}$ are hyperparameters used to balance the contribution of each component.}
The individual loss components are defined as follows:

\paragraph{1. Trajectory regression loss ($\ell_{\text{traj}}$).} This loss measures the $L_1$ error between the predicted 2D coordinates $\mathbf{M}_{2d}^{(t)}$ and the ground truth coordinates $\mathbf{M}_{2d}^{gt}$. Importantly, it is weighted by the ground truth visibility mask $\mathbf{V}^{gt}$: points that are visible in the target frame are assigned a full weight of 1, while points that are occluded or invisible receive a smaller weight of 0.2. This design encourages the model to focus more on accurately predicting visible points while still providing a weak supervision signal for invisible points:
\begin{equation}
\label{equ:traj}
\ell_{\text{traj}}^{(t)} =
\Big[ \mathds{1}(\mathbf{V}^{gt}) + 0.2 \cdot (1 - \mathds{1}(\mathbf{V}^{gt})) \Big]
\cdot 
\big\| \mathbf{M}_{2d}^{(t)} - \mathbf{M}_{2d}^{gt} \big\|_1.
\end{equation}

\paragraph{2. Visibility classification loss ($\ell_{\text{vis}}$).}
The visibility loss uses Binary Cross Entropy ($\text{BCE}$) to supervise the predicted visibility $\mathbf{V}'$, encouraging the model to correctly classify points as visible or occluded based on the ground truth mask $\mathbf{V}^{gt}$:

\begin{equation}
\ell_{\text{vis}}^{(t)} = \text{BCE}\left( \mathbf{V}'^{(t)}, \mathbf{V}^{gt} \right).
\end{equation}

\paragraph{3. Confidence loss ($\ell_{\text{conf}}$)}
Our model predicts a separate confidence score $\mathbf{V}''$ (track reliability or certainty), this loss supervises it using BCE. The ground truth label is a binary flag indicating whether the corresponding trajectory prediction falls within an acceptable pixel error tolerance $\delta$:

\begin{equation}
\ell_{\text{conf}}^{(t)} = \text{BCE}\left( \mathbf{V}''^{(t)}, \mathds{1}\left( \left\| \mathbf{M}_{2d}^{(t)} - \mathbf{M}_{2d}^{gt} \right\|_2 < \delta \right) \right).
\end{equation}

\subsection{Losses for 3D Branch}

For the 3D branch, we first apply the predicted 3D flow, $\mathbf{M}_{3d}^{(t)}$, to the first-frame camera-centric point cloud $\mathbf{P}_1$ to obtain the predicted corresponding points at all other frames, denoted as $\hat{\mathbf{P}}_{3d}^{(t)}$:
\begin{equation}
\hat{\mathbf{P}}_{3d}^{(t)} = \mathbf{P}_1 + \mathbf{M}_{3d}^{(t)}.
\end{equation}
The total loss for the 3D branch, $\mathcal{L}^{3d}_{\text{total}}$, is then composed of the 3D trajectory loss ($\mathcal{L}_{\text{traj}}^{3d}$) and the scene flow smoothness loss ($\mathcal{L}_{\text{smooth}}^{3d}$). Adhering to the iterative refinement structure of CoTracker~\cite{karaev2024cotracker}, the total loss is aggregated across all iterative steps $t$ using an exponential weighting factor $\gamma$:

\begin{equation}
\mathcal{L}^{3d}_{\text{total}} = \sum_{t=1}^{N_{iter}} \gamma^{N_{iter}-t} \left( \mathcal{L}_{\text{traj}}^{(t),3d} + \lambda_{\text{smooth}} \mathcal{L}_{\text{smooth}}^{(t),3d} \right),
\end{equation}
\noindent where $N_{iter}$ is the total number of update iterations and $\lambda_{\text{smooth}}$ is the hyperparameter balancing the contribution of the smoothness term.
\paragraph{1. Trajectory regression loss ($\ell_{\text{traj}}^{3d}$).} Subsequently, we supervise the full 3D trajectory by adopting a loss function structurally similar to the 2D trajectory loss (Equation \ref{equ:traj}). This 3D trajectory loss, $\ell_{\text{traj}}^{3d}$, computes the $L_1$ difference between the predicted 3D coordinates and the ground truth 3D coordinates, incorporating a robust weighting scheme based on ground truth visibility, $\mathbf{V}^{gt}$:

\begin{equation}
\ell_{\text{traj}}^{(t),3d}= \Big[ \mathds{1}(\mathbf{V}^{gt}) + 0.2 \cdot (1 - \mathds{1}(\mathbf{V}^{gt})) \Big]
\cdot 
\big\| \hat{\mathbf{P}}_{3d}^{(t)} - \mathbf{P}_{3d}^{gt} \big\|_1.
\end{equation}

\paragraph{2. Scene flow smoothness loss ($\ell_{\text{smooth}}^{3d}$).}

In addition to the trajectory supervision, we incorporate a scene flow smoothness loss($\ell_{\text{smooth}}^{3d}$) to enforce local rigidity and prevent erratic flow predictions across spatially adjacent points. Given the high dimensionality of the point cloud, calculating this loss across all $N$ points is computationally expensive ($\mathcal{O}(N^2)$). To address this, we adopt an efficient sampling strategy based on anchor points.

\begin{figure}[!h]
    \begin{center}
        \includegraphics[width=\linewidth]{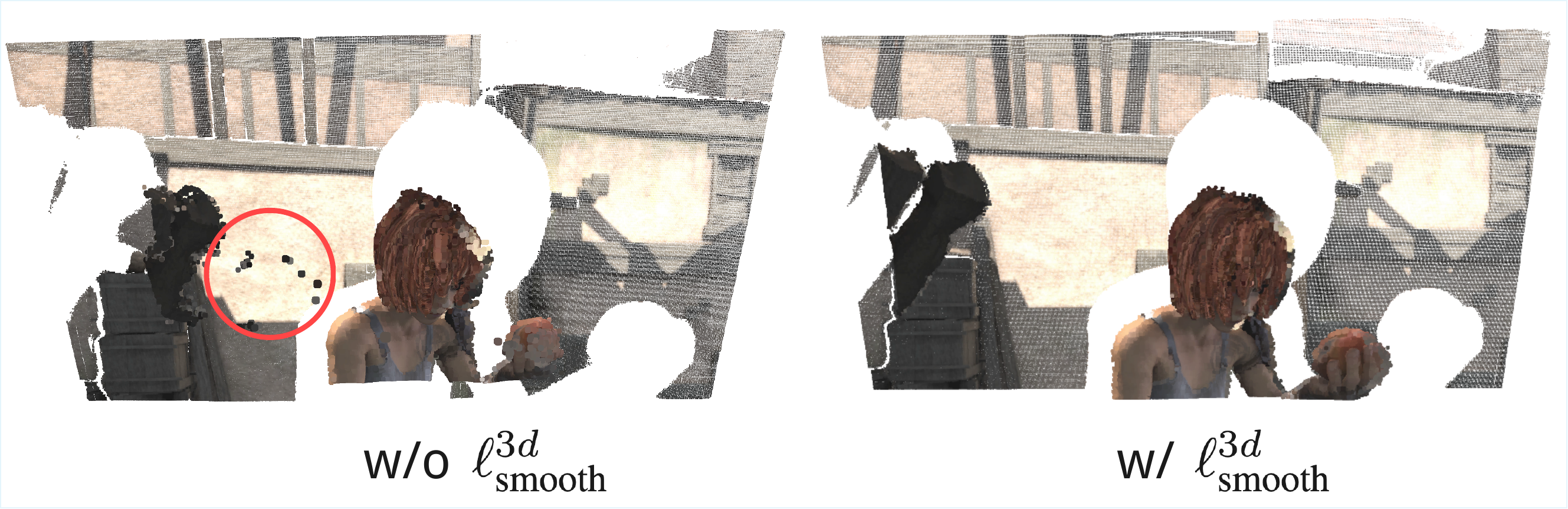}
\caption{\textbf{Effectiveness of $\ell_{\text{smooth}}^{3d}$ .}}
        \label{fig:sup_smoothloss}
    \end{center}
       \vspace{-1.5em}
\end{figure}

Specifically, we randomly sample $M$ anchor points ($\mathbf{P}_{anchor}$) from the ground-truth point cloud $\mathbf{P}_{3d}^{gt}$, where $M \ll N$. We then retrieve the corresponding $M$ predicted flow vectors ($\hat{\mathbf{M}}_{anchor}$) and use the spatial structure of $\mathbf{P}_{anchor}$ to establish local neighborhoods via $K$-Nearest Neighbors (KNN) search. The loss is computed as the squared $L_2$ distance between the flow vector of each anchor point and its $K$ neighbors within the sampled set:

\begin{equation}
\ell_{\text{smooth}}^{(t), 3d} = \frac{1}{M \cdot K} \sum_{i=1}^{M} \sum_{j \in \mathcal{N}_i^K} \left\| \hat{\mathbf{M}}^{(t)}_{anchor, i} - \hat{\mathbf{M}}^{(t)}_{anchor, j} \right\|_2^2,
\end{equation}
where $\hat{\mathbf{M}}^{(t)}_{anchor, i}$ is the predicted scene flow for the $i$-th anchor point at iteration $t$, and $\mathcal{N}_i^K$ denotes the set of $K$ nearest neighbors of point $i$ within the sampled $M$ points.  The effectiveness of this loss is illustrated in Fig.~\ref{fig:sup_smoothloss}, demonstrating its impact on improving scene flow accuracy.

\section{Test-time Pose Refinement}
\label{sec:pose_est}
We refine the per-frame camera poses \( \{\pi_t \in \mathrm{SE}(3)\} \) from the reconstructed 3D tracks. 
We begin by computing dynamic masks using VLM~\cite{achiam2023gpt} and Grounding-SAM~\cite{sam,liu2024grounding} to segment foreground dynamic objects. 
Static 3D tracks are then extracted by retaining only those tracks whose projections fall within the static regions.

Given these static tracks, we reproject each point from timestep \(t_1\) to timestep \(t_2\) under the current camera poses and optimize the poses using the following projection loss:
\begin{equation}
\begin{aligned}
\label{reprojection}
\ell_{\text{proj}}
&=
\sum_{i=1}^{N}
\sum_{t_1=1}^{T}
\sum_{t_2=1}^{T}
\left\|
\pi_{t_2}\pi_{t_1}^{-1}
\big(
\mathbf{P}_{\text{static}}(i,t_1)
\big)
\right. \\
&\qquad\qquad\left.
-
\mathbf{P}_{\text{static}}(i,t_2)
\right\|_2^2 ,
\end{aligned}
\end{equation}
where \(\pi_t(\cdot)\) denotes the transformation at timestep \(t\), 
\(\mathbf{P}_{\text{static}}(i,t)\in \mathbb{R}^3\) is the 3D location of the \(i\)-th static track at timestep \(t\), 
and \(N\) is the total number of static tracks.
To improve computational efficiency, we divide the entire video into \(C\) clips and estimate the camera poses of each clip in parallel. 
After optimizing the intra-clip poses, we estimate inter-clip transformations to merge all clips into a globally consistent camera trajectory. 
Finally, once the merged global poses are initialized, we refine all camera poses again by optimizing the reprojection loss in Eq.~\ref{reprojection} over the entire sequence. 
\section{Metric Details}
\label{sec:metrics}

We evaluate our method using a comprehensive set of metrics covering geometry estimation, motion estimation, and camera pose accuracy. Below, we denote the predicted value with $(\cdot)$ and the ground truth with $\hat{(\cdot)}$. The set of valid pixels or points is denoted by $\Omega$, and $N = |\Omega|$.

\subsection{Depth and Geometry Metrics}
\noindent \textbf{Abs Rel (Absolute Relative Error):}
This metric measures the mean absolute relative difference between the predicted depth $d_i$ and the ground truth depth $\hat{d}_i$. It is defined as:
\begin{equation}
    \text{Abs Rel} = \frac{1}{N} \sum_{i \in \Omega} \frac{|\hat{d}_i - d_i|}{\hat{d}_i}.
\end{equation}

\noindent \textbf{$\boldsymbol{\delta < 1.25}$ (Threshold Accuracy):}
This metric measures the percentage of pixels where the ratio between the predicted and ground truth depth is within a threshold of $1.25$. It indicates the prevalence of accurate predictions:
\begin{equation}
    \delta < \tau = \frac{1}{N} \sum_{i \in \Omega} \mathbb{I}\left( \max\left(\frac{\hat{d}_i}{d_i}, \frac{d_i}{\hat{d}_i}\right) < \tau \right),
\end{equation}
where $\tau = 1.25$, and $\mathbb{I}(\cdot)$ is the indicator function.

\subsection{3D Scene Flow Metrics}
\noindent \textbf{EPE3D (3D End-Point Error):}
EPE3D measures the average Euclidean distance (in meters) between the predicted 3D scene flow vectors and the ground truth:
\begin{equation}
    \text{EPE3D} = \frac{1}{N} \sum_{i \in \Omega} \| \mathbf{p}_i - \hat{\mathbf{p}}_i \|_2,
\end{equation}
where $\mathbf{p}_i$ is the predicted flow vector and $\hat{\mathbf{p}}_i$ is the ground truth.

\noindent \textbf{AccS (Strict Accuracy - 3D):}
The percentage of points whose EPE3D is strictly within a tight threshold. Following standard scene flow protocols, a point is considered accurate if:
\begin{equation}
    \text{AccS} = \frac{1}{N} \sum_{i \in \Omega} \mathbb{I}\left( \| \mathbf{p}_i - \hat{\mathbf{p}}_i \|_2 < \tau_1 \lor \frac{\| \mathbf{p}_i - \hat{\mathbf{p}}_i \|_2}{\| \hat{\mathbf{p}}_i \|_2} < \tau_2 \right).
\end{equation}
Typically, $\tau_1 = 0.05$m and $\tau_2 = 5\%$. Note that the relative error is normalized by the ground truth magnitude $\|\hat{\mathbf{p}}_i\|_2$.

\noindent \textbf{AccR (Relaxed Accuracy - 3D):}
Similar to AccS but with relaxed thresholds (typically $\tau_1 = 0.10$m and $\tau_2 = 10\%$) to evaluate robustness against larger motions.

\subsection{2D Optical Flow Metrics}
\noindent \textbf{EPE2D (2D End-Point Error):}
The average Euclidean distance (in pixels) between the predicted 2D coordinates $\mathbf{u}_i$ and ground truth $\hat{\mathbf{u}}_i$:
\begin{equation}
    \text{EPE2D} = \frac{1}{N} \sum_{i \in \Omega} \| \mathbf{u}_i - \hat{\mathbf{u}}_i \|_2.
\end{equation}

\noindent \textbf{AccS2D (Strict Accuracy - 2D):}
The percentage of points with EPE2D smaller than a strict threshold (1 pixel):
\begin{equation}
    \text{AccS2D} = \frac{1}{N} \sum_{i \in \Omega} \mathbb{I}\left( \| \mathbf{u}_i - \hat{\mathbf{u}}_i \|_2 < \tau_{\text{strict}} \right).
\end{equation}

\noindent \textbf{AccR2D (Relaxed Accuracy - 2D):}
The percentage of points satisfying a looser threshold $\tau_{\text{relaxed}}$ (3 pixels) to measure coarse tracking capability.

\subsection{Long-term Point Tracking Metrics (TAP-Vid)}
\noindent \textbf{APD (Average Position Deviation):}
This metric measures the percentage of points whose predicted positions fall within a specified threshold relative to the ground-truth depth.

\noindent \textbf{AJ (Average Jaccard):}
Also known as the "Position Accuracy" in some benchmarks, it measures the fraction of points that are within a specified distance threshold from the ground truth, averaged over the sequence. It is essentially a survival rate weighted by spatial accuracy.

\noindent \textbf{$\boldsymbol{\delta_{\text{avg}}^{\text{vis}}}$ (Average Visible Delta):}
This metric specifically measures the position accuracy for points that are currently visible (not occluded). It computes the average percentage of visible points where the prediction error is below a specific threshold $\delta$.

\noindent \textbf{OA (Occlusion Accuracy):}
OA evaluates the binary classification performance of the visibility estimation head. It is the accuracy of predicting whether a point is occluded or visible:
\begin{equation}
    \text{OA} = \frac{1}{N_{\text{total}}} \sum_{i} \mathbb{I}(v_i = \hat{v}_i),
\end{equation}
where $\hat{v}_i \in \{0, 1\}$ denotes the ground-truth visibility status and $v_i$ denotes the prediction.

\subsection{Camera Pose Metrics}
\noindent \textbf{ATE (Absolute Trajectory Error):}
ATE measures the global consistency of the estimated camera trajectory. It computes the root-mean-square error (RMSE) of the translation difference between the estimated trajectory $\mathbf{T}_{est}$ and ground truth $\mathbf{T}_{gt}$ after aligning them via a similarity transformation (Sim3) or rigid transformation (SE3):
\begin{equation}
    \text{ATE} = \sqrt{ \frac{1}{M} \sum_{j=1}^{M} \| \text{trans}(\mathbf{T}_{gt,j}^{-1} \mathbf{S} \mathbf{T}_{est,j}) \|_2^2 }.
\end{equation}

\noindent \textbf{RTE (Relative Trajectory Error):}
RTE measures local consistency by averaging the drift over fixed time intervals or distances. It evaluates the error in the relative pose between pairs of frames separated by a fixed lag.

\noindent \textbf{RRE (Relative Rotation Error):}
RRE measures the geodesic distance between the predicted and ground truth rotation matrices, averaged over all frame pairs. It is defined as:
\begin{equation}
    \text{RRE} = \frac{1}{M} \sum_{j} \arccos \left( \frac{\text{trace}(\mathbf{R}_{gt,j}^T \mathbf{R}_{est,j}) - 1}{2} \right).
\end{equation}

\section{More Ablation Study}
\label{sec:more_ablation}
\paragraph{Geometry encoder.}
We provide additional ablation experiments on the geometry encoder, as summarized in Tab.~\ref{table:geo}. This ablation is conducted using MoGe~\cite{wang2025moge1} as the geometry backbone initialization. Replacing our affine-invariant loss with a scale-invariant one (``w/ scale-inv'') degrades performance, demonstrating the effectiveness of the affine-invariant formulation. Removing the regularization loss ($\ell_{reg}$) leads to a catastrophic drop (Abs Rel 0.3530), highlighting its crucial role in stabilizing convergence. Finally, removing the local loss ($\ell_{local}$) also results in a slight degradation (Abs Rel 0.1053), confirming its contribution to refining geometric details.

\label{ablation}
\begin{table}[!ht]
  \centering
\vspace{-0.8em}
  \footnotesize
  \setlength\tabcolsep{5pt}
  \renewcommand{\arraystretch}{1.1}
  \caption{\textbf{Ablation study on geometry encoder.} The results are averaged over all datasets. \textbf{Best} results are highlighted in \colorbox{best}{darker blue}, and \textbf{second best} in \colorbox{second}{lighter blue}.}
  \label{table:geo}
  \begin{tabular}{lcc}
    \toprule
    \textbf{Setting} & Abs Rel $\downarrow$ & $\delta < 1.25 \uparrow$ \\
    \midrule
    w/ scale-inv  & 0.1060 & 0.8788 \\
    w/o $\ell_{reg}$ & 0.3530 & 0.3547 \\
    w/o $\ell_{local}$ &  \cellcolor{second}{0.1053} &  \cellcolor{second}{0.8790}  \\
    \midrule
    \textbf{Full (Ours)} & \cellcolor{best}{0.0973} &\cellcolor{best}{0.8921}\\
    \bottomrule
  \end{tabular}
  \vspace{-0.8em}
\end{table}

\begin{table}[t]
\centering
\caption{\textbf{Efficiency comparison on 16-frame ADT~\cite{pan2023aria} sequences.} Gray rows indicate Out-of-Memory (OOM) failures under dense tracking settings. \colorbox{best}{Best} and \colorbox{second}{second best} results are highlighted.}
\label{tab:efficiency}
\begin{adjustbox}{width=0.8\textwidth}
\begin{tabular}{
    lcccc
}
\toprule
Method & ATE$\downarrow$ & RTE$\downarrow$ & RRE$\downarrow$ & Time (s) \\
\midrule
STV2~\cite{xiao2025spatialtrackerv2} (FF.)
& 0.133 & 0.057 & 0.641 &  \cellcolor{second}{9} \\
DA3~\cite{lin2025depth}(FF.)
& 0.124 & 0.061 & 0.331 & \cellcolor{best}{6} \\
\textbf{Ours (FF.)}
& 0.119 & 0.054 & 0.309 &\cellcolor{best}{6} \\
\midrule
MegaSaM~\cite{li2024megasam} (Opt.)
&  \cellcolor{second}{0.059} &  \cellcolor{second}{0.027} &  \cellcolor{second}{0.120} & 310 \\
\textbf{Ours (FF.+Opt.)}
& \cellcolor{best}{0.045} & \cellcolor{best}{0.019} & \cellcolor{best}{0.115} & 294 \\
\bottomrule
\end{tabular}
\end{adjustbox}
\end{table}
\noindent \textbf{Test-time pose refinement.}
Leveraging long-term correspondences, predicted point clouds, and initial poses, our method supports test-time pose refinement via Bundle Adjustment.
Tab.~\ref{tab:efficiency} evaluates camera pose accuracy on the Sintel~\cite{butler2012naturalistic} dataset, where ``Ours (FF.)'' consistently outperforms the geometry-and-motion-based method  STV2. Applying our optimization strategy (detailed in Supp.~Sec.~E) refines these predictions further; this optimized variant, ``Ours (FF.+Opt.)'', achieves higher accuracy than MegaSaM~\cite{li2024megasam} at a lower computational cost.

\begin{figure}[!h]
    \vspace{-1em}
    \begin{center}
        \includegraphics[width=\linewidth]{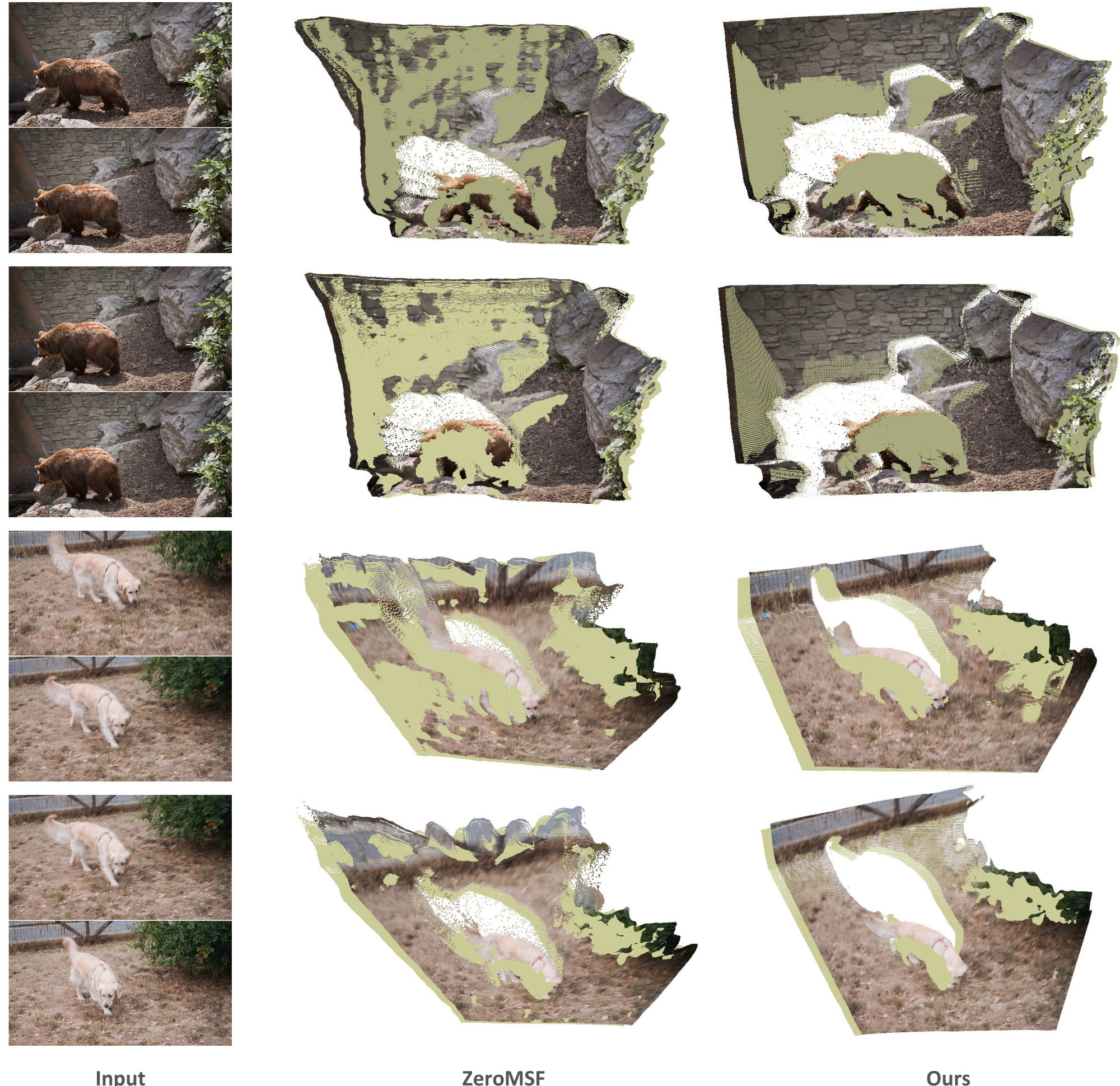}
\caption{\textbf{Scene flow Visualization.} The deformed point maps (colored \textcolor{deformedcolor}{\rule{1em}{1em}}) show that our method produces more temporally consistent geometry and motion.}
        \label{fig:compare}
    \end{center}
    \vspace{-1.5em}
\end{figure}
\section{More Visualization}
\label{sec:more_vis}
\begin{figure*}[!h]
    \vspace{-1em}
    \begin{center}
        \includegraphics[width=\linewidth]{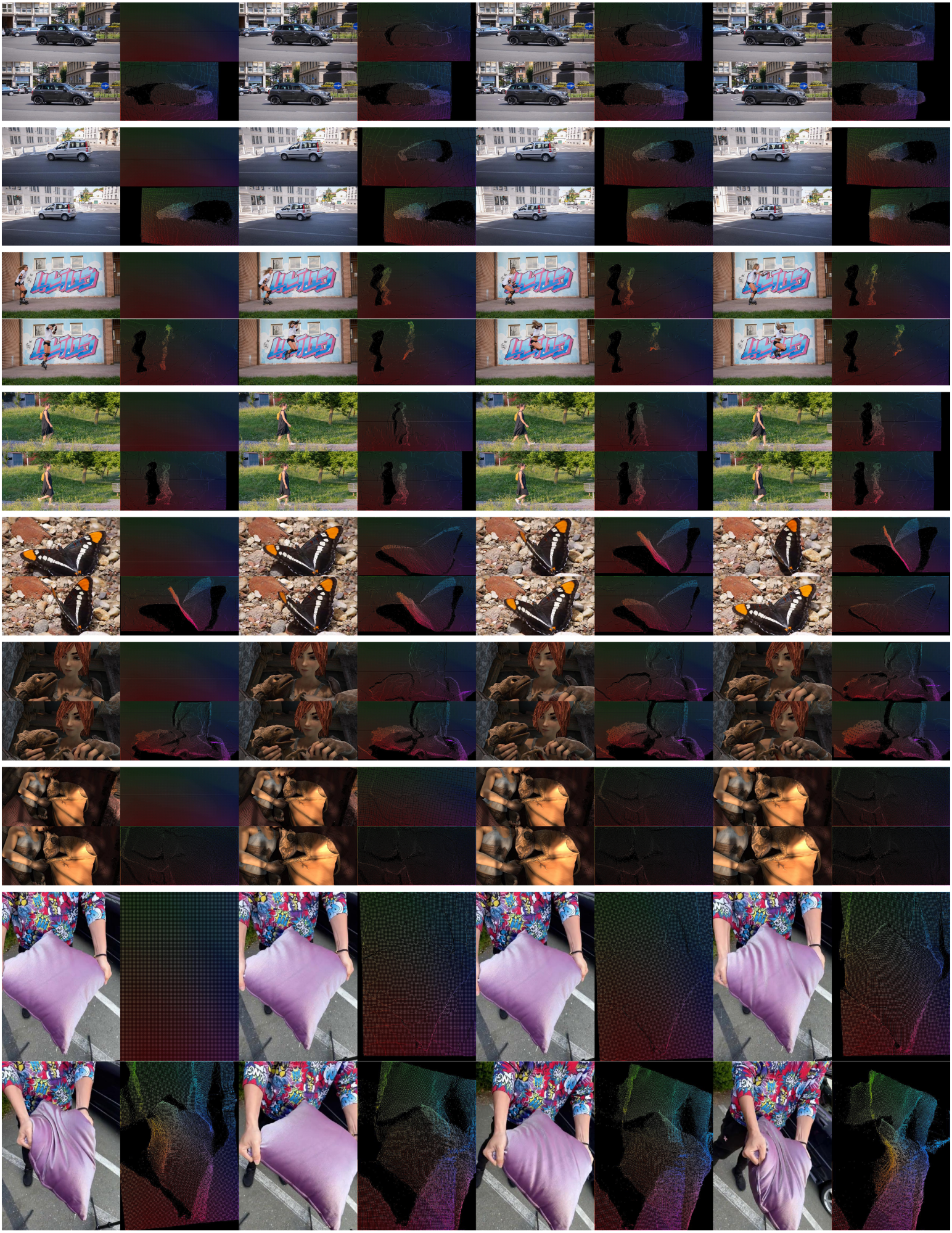}
\caption{\textbf{Visualization of first-frame 2D dense tracking.}}
        \label{fig:2dvis}
    \end{center}
    \vspace{-1.5em}
\end{figure*}

\begin{figure*}[!h]
    \vspace{-1em}
    \begin{center}
        \includegraphics[width=\linewidth]{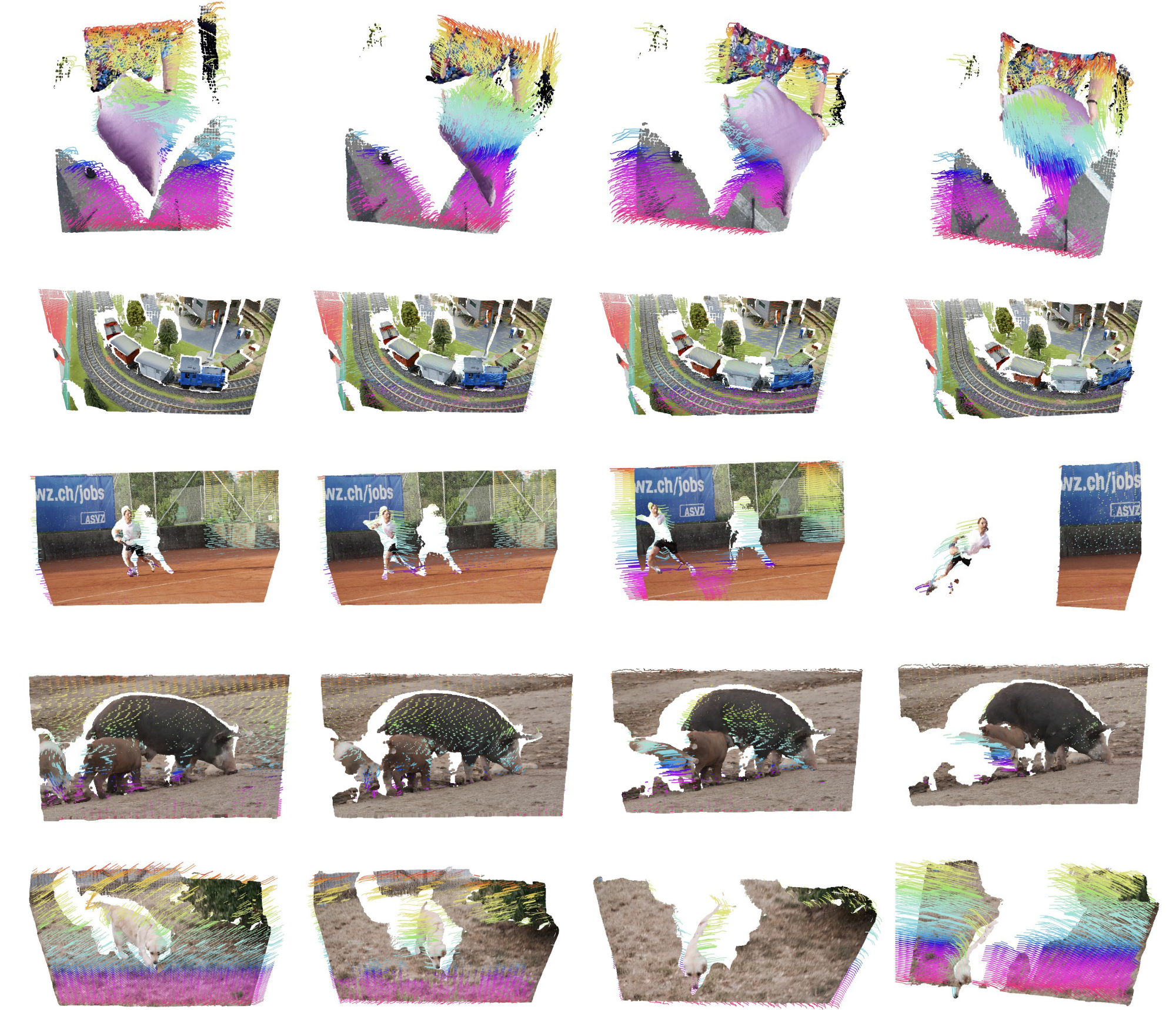}
\caption{\textbf{Visualization of first-frame 3D dense tracking.}}
        \label{fig:3dvis}
    \end{center}
    \vspace{-1.5em}
\end{figure*}

\begin{figure*}[!h]
    \vspace{-1em}
    \begin{center}
        \includegraphics[width=\linewidth]{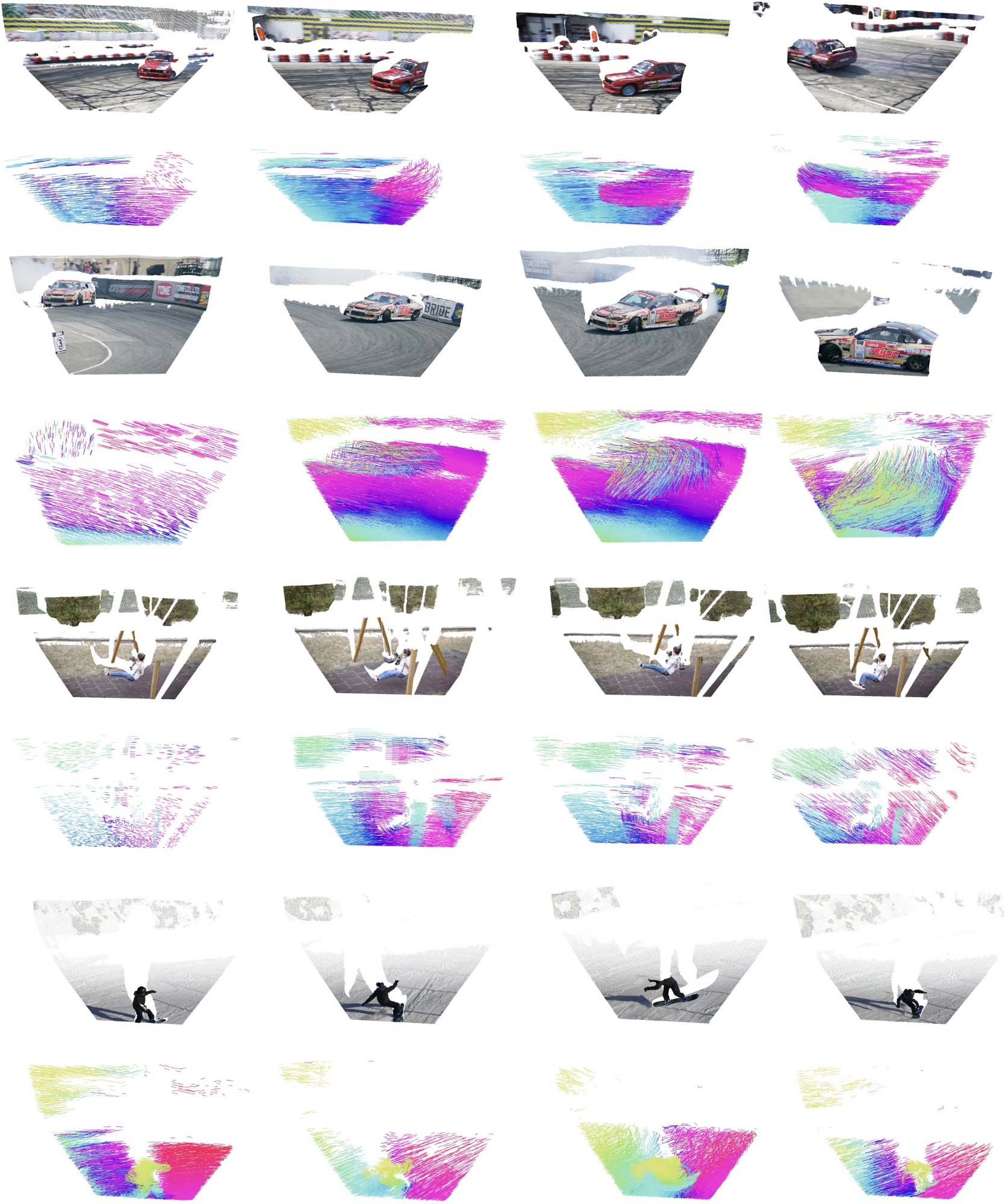}
\caption{\textbf{Visualization of dense per-pixel trajectories across all frames.}}
        \label{fig:3dvisall}
    \end{center}
    \vspace{-1.5em}
\end{figure*}

\begin{figure*}[!h]
    \vspace{-1em}
    \begin{center}
        \includegraphics[width=0.9\linewidth]{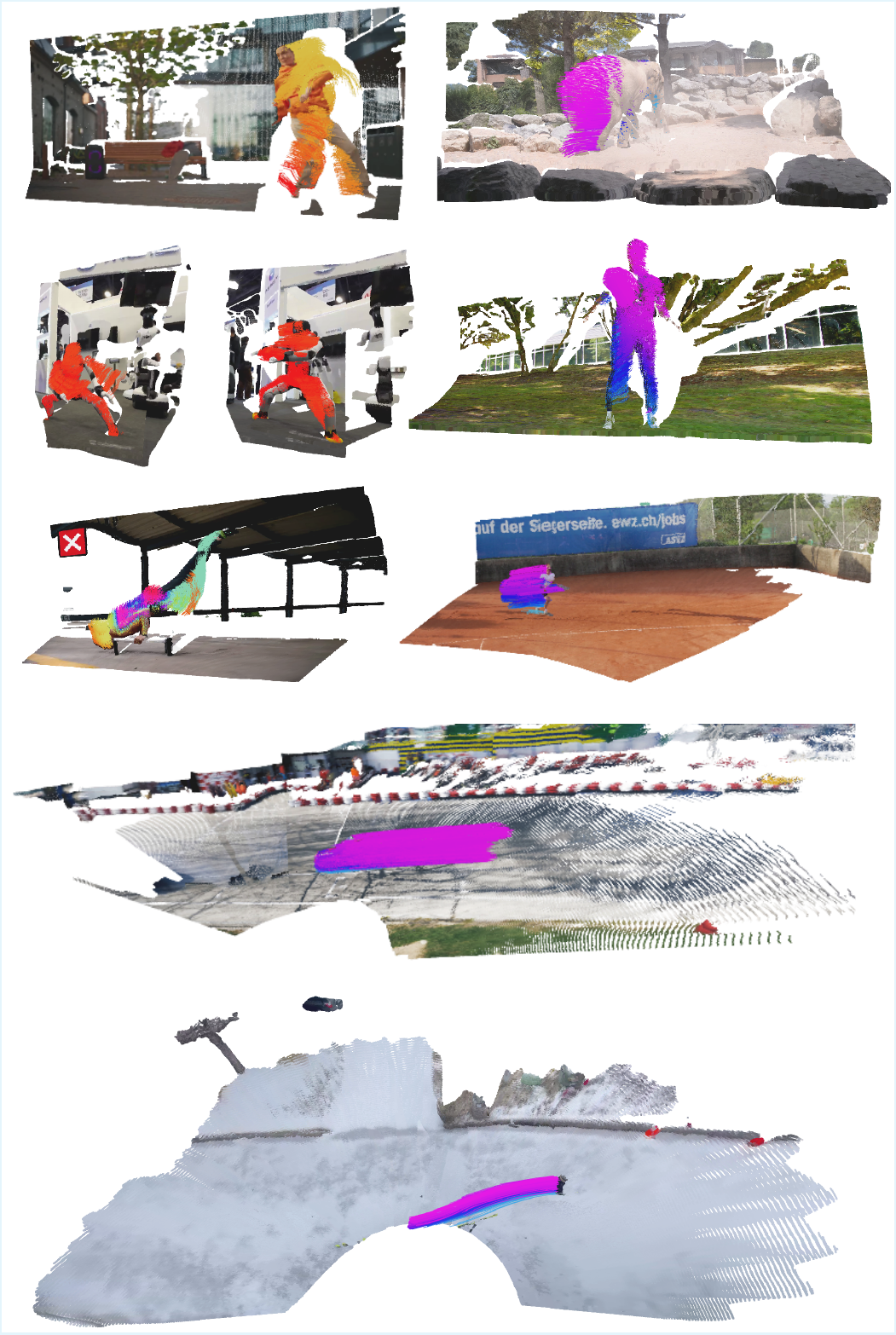}
\caption{\textbf{Visualization of the dense trajectories across all frames in the global world coordinate system.}}
        \label{fig:3dvisworld}
    \end{center}
    \vspace{-1.5em}
\end{figure*}

To further demonstrate the capability of our framework, we provide additional qualitative results. Specifically, we present camera-centric results, including scene flow estimation, dense first-frame 2D tracking, dense first-frame 3D tracking, and per-pixel tracking across all frames (Fig.~\ref{fig:compare}, Fig.~\ref{fig:2dvis}, Fig.~\ref{fig:3dvis}, and Fig.~\ref{fig:3dvisall}). We also show world-centric dense tracking across all frames in Fig.~\ref{fig:3dvisworld}.
(1) \textbf{Scene Flow Visualization.}  
We first present pairwise scene flow predictions between two frames. Scene flow characterizes the 3D motion of each point and serves as an essential cue for understanding dynamic geometries. In our visualization, the deformed point maps (colored \textcolor{deformedcolor}{\rule{1em}{1em}}) highlight the predicted per-point displacements. Compared with the current state-of-the-art method ZeroMSF~\cite{liang2025zero}, our results exhibit smoother spatial transitions and significantly higher temporal consistency in both geometry and motion. This demonstrates that our deformation field models coherent 3D dynamics more effectively, even in complex and rapidly changing scenes.
(2) \textbf{Dense 2D Tracking from the First Frame.}  
The second group shows dense 2D correspondences anchored at the first frame. Pixels with the same color trace the evolution of the same point across time, revealing long-term and fine-grained motion patterns throughout the sequence.
(3) \textbf{Dense 3D Trajectories from the First Frame.}  
We further visualize reconstructed 3D trajectories originating from the first frame. For clarity and rendering efficiency, we uniformly subsample the trajectories while preserving their overall motion structures and temporal smoothness.
(4) \textbf{Dense Per-Pixel Trajectories Across All Frames.}  
We present dense per-pixel trajectories over the full sequence (second row), which consistently track motions even when new objects enter the scene, highlighting the robustness of our method in handling complex dynamic scenarios.
(5) \textbf{Dense Trajectories in the World Coordinate System.} Finally, we visualize dense trajectories across all frames after transforming them into a unified global world coordinate system. This world-centric representation effectively decouples camera ego-motion from object dynamics: static background elements remain spatially stable over time, while dynamic objects exhibit coherent and physically meaningful absolute motion trajectories in 3D space.
For more intuitive demonstrations and continuous motion effects, we refer the reader to the supplementary video.
\section{Discussion with Concurrent work}
\begin{figure}[!h]
    \vspace{-1em}
    \begin{center}
        \includegraphics[width=\linewidth]{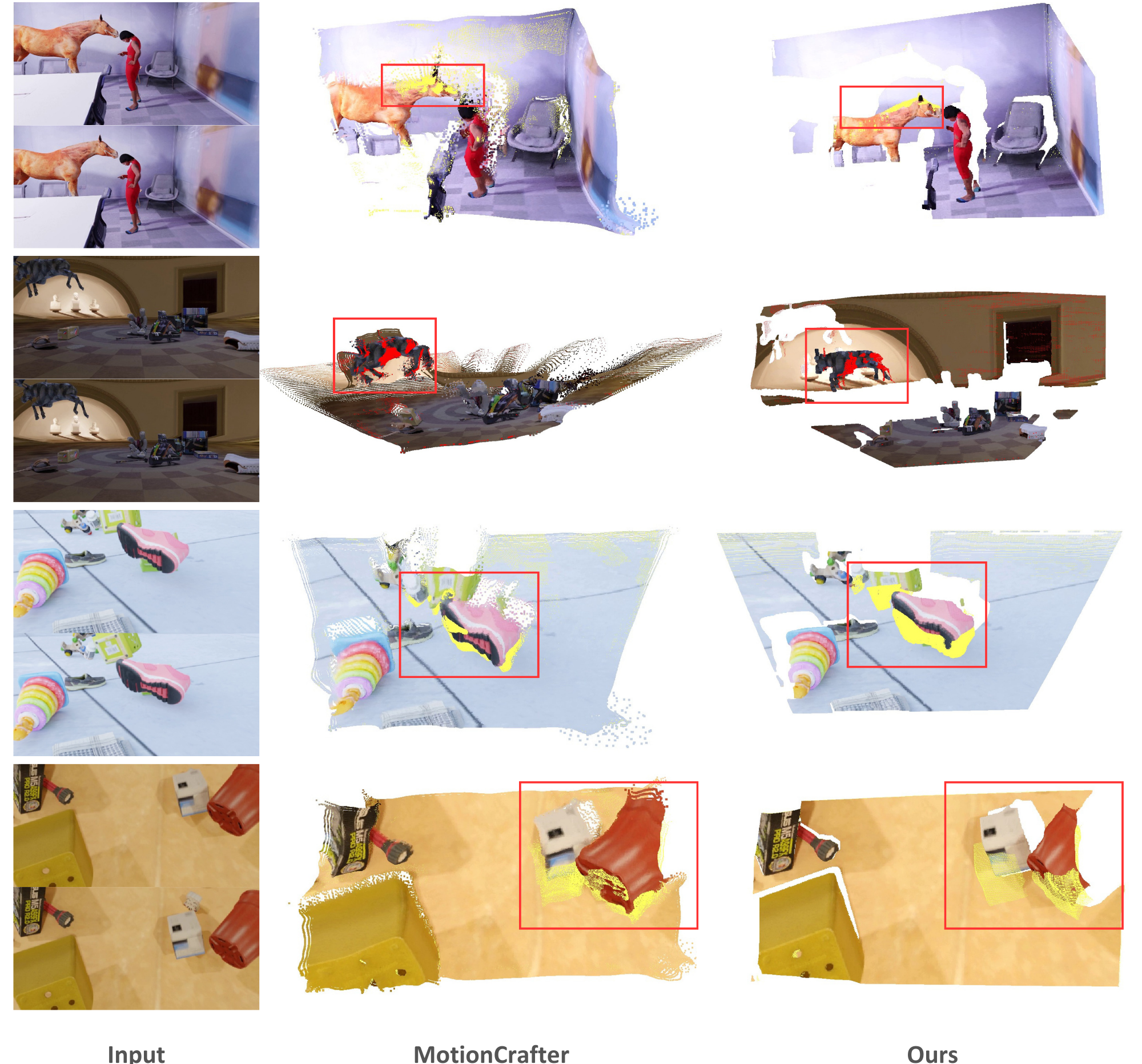}
\caption{\textbf{Scene Flow Visualization.} The deformed point maps (shown in solid colors) indicate that our method produces higher-quality geometry and motion.}
        \label{fig:compare_motioncrafter}
    \end{center}
    \vspace{-1.5em}
\end{figure}

MotionCrafter~\cite{zhu2026motioncrafterdensegeometrymotion} is our concurrent work that also investigates joint reconstruction of geometry and motion.
The key difference lies in the modeling paradigm: MotionCrafter explores the potential of diffusion models for joint geometry–motion reconstruction, whereas our method adopts a purely ViT-based feed-forward framework.
MotionCrafter primarily focuses on long-sequence motion modeling by predicting motion between consecutive frames (i.e., frame $1 \rightarrow 2$, $2 \rightarrow 3$, $\ldots$, $N-1 \rightarrow N$).
In contrast, our method is more flexible, as it supports on-demand arbitrary-pair motion prediction, enabling consistent 4D reasoning across non-consecutive frames.
We compare the world-centric geometric reconstruction results in Tab.~\ref{tab:world-centric}, and provide qualitative scene flow comparisons in Fig.~\ref{fig:compare_motioncrafter}.

\begin{table}[!ht]
\centering
\caption{\textbf{Evaluation on world-centric geometric reconstruction.}}
\label{tab:world-centric}
\footnotesize
\setlength\tabcolsep{5pt}
\renewcommand{\arraystretch}{1.1}
\begin{adjustbox}{width=1\textwidth}\begin{tabular}{lcccccccc}
\toprule
Method 
& \multicolumn{2}{c}{Sintel} 
& \multicolumn{2}{c}{GMUKitchen} 
& \multicolumn{2}{c}{Monkaa} 
& \multicolumn{2}{c}{Scannet test} \\
\cmidrule(lr){2-3} \cmidrule(lr){4-5} \cmidrule(lr){6-7} \cmidrule(lr){8-9}
 & Abs Rel $\downarrow$ & $\delta < 1.25 \uparrow$
 & Abs Rel $\downarrow$ & $\delta < 1.25 \uparrow$
 & Abs Rel $\downarrow$ & $\delta < 1.25 \uparrow$ 
 & Abs Rel $\downarrow$ & $\delta < 1.25 \uparrow$\\
\midrule

MotionCrafter~\cite{zhu2026motioncrafterdensegeometrymotion}
& 0.2192 & 0.6942
& 0.1303 & 0.8991
& 0.1864 & 0.7837 &0.1055 &0.9663 \\
\textbf{Ours}
& \textbf{0.1408} & \textbf{0.8155}
& \textbf{0.0397} & \textbf{0.9588}
& \textbf{0.1503} & \textbf{0.8083} & \textbf{0.0237}&  \textbf{0.9931} \\
\bottomrule
\end{tabular}
\end{adjustbox}
\end{table}

\section{Limitations and Future Work}
\label{sec:limitations}

Despite the effectiveness of our proposed approach, several limitations remain. A primary constraint is the dependence on captured 4D motion datasets, which are labor-intensive to acquire and limited in scale. Consequently, the model may struggle to generalize to extreme poses or complex topological changes not represented in the training set.
In the future, we plan to explore synthetic data generation pipelines, potentially utilizing generative diffusion models or physics engines, to create large-scale and diverse training samples. Furthermore, we aim to investigate unsupervised or semi-supervised learning schemes to reduce the dependency on labeled data, ultimately pushing the boundaries of 4D reconstruction in generalizable settings.


{
    \small
    \bibliographystyle{ieeenat_fullname}
    \bibliography{main}
}


\end{document}